\documentclass[a4paper,fleqn]{cas-sc}
\usepackage{float}
\usepackage[numbers]{natbib}
\usepackage{algorithm}
\usepackage{algorithmic}
\usepackage{placeins}
\graphicspath{{images/}}

\newcommand{\hl}[1]{#1}
\renewcommand{\printorcid}{}

\begin{document}
\let\WriteBookmarks\relax

\shorttitle{RLDE: A Reinforcement Learning Framework for Interpretable Iterative Document Editing}
\shortauthors{Nikhil Parkar et~al.}

\title [mode = title]{RLDE: A Reinforcement Learning Framework for Interpretable Iterative Document Editing}

\author[1]{Nikhil Parkar}
\ead{nikhil.parkar2023@vitstudent.ac.in}

\author[1]{M. Marimuthu}
\cormark[1]
\ead{marimuthu.m@vit.ac.in}

\author[1]{Riya Joshi}
\ead{riyarahul.joshi2023@vitstudent.ac.in}

\author[1]{Karthikeyan N}
\ead{n.karthikeyan@vit.ac.in}

\address[1]{School of Computer Science and Engineering, Vellore Institute of Technology, Chennai Campus, Chennai 600127, Tamil Nadu, India}

\cortext[cor1]{Corresponding author}

\begin{abstract}
Improving the quality, clarity, and flow of long documents remains a challenging problem in NLP, as human editing typically occurs through iterative refinement rather than a single-pass rewrite. Most existing automated systems act as black-box generators or rely on supervised learning, limiting interpretability and failing to capture the sequential nature of revisions. In this paper, we propose RLDE, an automated end-to-end metric-driven RL framework that formulates document revision as a reference-free MDP problem (unsupervised with respect to paired document-revision targets). Working with limited action space (\textit{NOOP}, \textit{INSERT}, \textit{DELETE}, \textit{REPLACE}), RLDE leverages Pointer Networks for dynamic span selection on the token level and sends generative actions to the frozen BART-base model to get context-aware suggestions. PPO agents learn to optimize a multi-objective reward function based on readability, perplexity, and grammar metrics with the help of a novel SBERT-based Semantic Preservation Penalty constraint to avoid deletion degeneracy. Evaluations on the BEA-2019 dataset showed significant statistical results ($p< 0.01$) according to all the metrics where Flesch Reading Ease increased from 49.8 to 59.4, perplexity decreased from 63.7 to 39.8, while preserving $93\%$ of semantic information. When compared to Chain-of-Thought (CoT) prompt-engineered models like GPT-4 and LLaMA-2, RLDE shows high performance at a very low computational inference cost. Moreover, the quality of edits generated by RLDE has been verified via human evaluations scoring 4.9/5 for meaning preservation and 4.4/5 for edit quality.

\end{abstract}

\maketitle

\section{Introduction}
Improving the quality of long documents in terms of clarity, coherence, and language quality is still a challenge in Natural Language Processing (NLP). Although large language models have made improvements in text generation and grammatical error correction, automated document editing systems are still unable to capture the iterative process of human editing \cite{du2022acl}. Editors tend to undertake several rounds of revisions of the document rather than performing everything from scratch. However, modeling this sequential process of refinement is still an open problem in NLP research \cite{dong2019acl}.
Currently, there are two types of automated editing systems. The first type is rule-based systems that use manually designed linguistic edits \cite{anoushka2025}. However, these algorithms are inflexible and inefficient in processing higher-level features such as readability and document flow. The second type of systems is supervised learning models and large language models (LLMs), which usually perform single-pass text rewriting based on paired before-and-after examples or through prompt-based text rewriting \cite{yuan2016neural}. Although these systems are useful for localized text editing, they do not model the process of text revision as a sequence of structured edits.

In order to handle these difficulties, RLDE (Reinforcement Learning based Document Editing) technique is proposed, which considers the problem of document editing as a sequence decision making problem. In RLDE, a document is represented as an editable environment, and an agent interacts with it over several timesteps by choosing one of four discrete editing actions: NOOP, INSERT, DELETE, or REPLACE.

The agent is trained through the Proximal Policy Optimization (PPO) algorithm \cite{schulman2017ppo}, and is guided through a multi-objective reward function that takes into consideration the paragraph-level gains. The reward function combines gains in readability (Flesch Reading Ease), perplexity, and grammar scores \cite{yanamoto2022}. By using feedback that is driven by metrics, RLDE is able to learn editing policies from evaluative feedback without requiring labeled pairs of revisions.

\hl{While prior automated editing systems operate at the single-sentence level, depend on paired supervised corpora, or generate text through opaque black-box LLMs, this paper proposes RLDE, a} \textbf{\hl{reference-free, constrained multi-objective Markov Decision Process (MDP)}} \hl{for document-level refinement. While RLDE utilizes established underlying components (PPO, SBERT, Pointer Networks, and BART-base), its substantive methodological advance does not lie in a naive assembly of tools, but rather in: (i) formulating document editing as a reference-free sequential optimization problem (unsupervised with respect to paired document-revision targets) governed by intrinsic quality rewards without requiring any paired revision corpora; (ii) resolving the persistent ``catastrophic deletion'' and reward-hacking collapse of RL simplification through SBERT-based semantic preservation and length constraints; and (iii) decoupling discrete structural control from continuous language generation to achieve full decision-level interpretability.}

The primary objectives and contributions of this research are as follows:

\begin{itemize}
    \item \textbf{Reference-Free Document MDP}: To formulate document-level iterative text revision as a reference-free sequential decision process (unsupervised with respect to paired document-revision targets) guided by intrinsic linguistic feedback, eliminating the reliance on expensive paired correction corpora.
    \item \textbf{Constrained Multi-Objective Reward}: To design a theoretically grounded reward function that balances readability, fluency, and grammatical correctness while preventing deletion degeneracy via semantic preservation and length penalties.
    \item \textbf{Decoupled Control-Generation Architecture}: To decouple discrete structural decision-making (Pointer Network span selection over atomic primitives) from continuous text generation (frozen BART-base prior), ensuring full step-by-step decision interpretability.
    \item \textbf{Coarse-to-Fine Hierarchical Scoping}: To structure document refinement through a two-tier Manager-Worker framework, scaling reinforcement learning from paragraph selection down to sentence-level token edits with verified policy stability.
\end{itemize}

Overall, RLDE offers an interpretable, controllable, and computationally efficient framework for iterative refinement of documents, closely aligned with human revision.
The remainder of this paper is organized as follows. Section 2 reviews related work in text editing and RL, establishing the precise research gap RLDE fills. Section 3 presents the proposed methodology and MDP formulation. Section 4 reports experimental results, human evaluations, and ablation studies. Section 5 concludes the paper, and Section 6 outlines future research directions.

 \section{Literature Review}

The evolution of automated text editing spans sequence-to-sequence neural rewriting, non-autoregressive edit-tagging, imitation learning, and reinforcement learning (RL). Early methods treated editing as monotone sequence transduction, which suffered from severe exposure bias and single-pass irreversibility. To overcome these constraints, modern paradigms have branched into iterative non-autoregressive (NAR) generation, edit-based tagging, and RL-guided decoding. However, as summarized in Table~\ref{tab:editing_comparison}, existing paradigms exhibit critical methodological limitations when applied to document-level refinement.

\begin{table*}[t]
\centering
\caption{Comparative Analysis of Text Editing Paradigms, Controllable RL Frameworks, and RLDE}
\label{tab:editing_comparison}
\footnotesize
\renewcommand{\arraystretch}{1.22}
\setlength{\tabcolsep}{3pt}
\begin{tabular*}{\textwidth}{@{\extracolsep{\fill}}p{2.1cm}p{2.5cm}p{1.9cm}cp{2.2cm}cp{3.2cm}@{}}
\toprule
\textbf{Approach / Model} & \textbf{Editing Paradigm} & \textbf{Supervision / Data} & \textbf{Iterative} & \textbf{Interpretability} & \textbf{Scope} & \textbf{Key Limitation / Research Gap} \\
\midrule
GECToR \cite{omelianchuk2020gector} & Sequence tagging & Supervised (paired) & Greedy & Explicit token tags & Sentence & Cannot optimize non-differentiable quality metrics (readability, flow). \\
Levenshtein Tr. \cite{gu2019} & Non-autoregressive & Supervised (imitation) & Yes & Insert/Delete ops & Sentence & Exposure bias; lacks reward-driven multi-objective feedback. \\
Felix \cite{mallinson2020improve} & Tagging + insertion & Supervised (paired) & 2-pass & Pointer + tag ops & Sentence & Strictly supervised; no intrinsic reward-driven quality search. \\
Du et al. \cite{du2022acl} & Human revision models & Supervised (edit logs) & Yes & Revision categories & Doc & Descriptive/supervised; lacks autonomous policy optimization. \\
Agrawal et al. \cite{agrawal2022} & Imitation curriculum & Supervised / imitation & Yes & Span replacement & Sentence & Distribution drift; cannot balance conflicting quality objectives. \\
Yanamoto et al. \cite{yanamoto2022} & RL text simplification & Unsupervised (SARI) & Few-step & Continuous policy & Sentence & \textbf{Catastrophic deletion}: erases text to cheat readability; no semantic bounds. \\
Wang et al. \cite{wang2024} & Edit-based RL (NMT) & Reference BLEU & Yes & Discrete edit ops & Sentence & Machine translation only; requires gold reference strings for rewards. \\
Chen et al. (MEC) \cite{chen2024} & Fine-grained RL (LLM) & Supervised reward & Single & Black-box tokens & Sentence & Opaque token generation; heavy compute; lacks explicit edit trace. \\
CoT LLMs (GPT-4) \cite{stiennon2020summarization} & Prompted generation & Pretrained (zero-shot) & Single & Black box & Doc & Hallucination risk; rewrites entire text; high latency; unconstrained. \\
\midrule
\textbf{RLDE (Proposed)} & \textbf{Constrained MDP + Ptr-Net + BART} & \textbf{Reference-free (Metric deltas)} & \textbf{Yes} & \textbf{High (Primitives + Reward Deltas)} & \textbf{Doc} & \textbf{Fills gap}: autonomous multi-metric revision with semantic preservation bounds. \\
\bottomrule
\end{tabular*}
\end{table*}

\subsection{Grammatical Error Correction (GEC) \& Text Simplification}
Foundational GEC methods cast correction as sequence-to-sequence translation using large parallel corpora \cite{yuan2016neural}. While effective, autoregressive seq2seq models suffer from high inference latency and frequently overwrite valid context. Modern token-tagging models such as GECToR \cite{omelianchuk2020gector} mitigate this by predicting discrete token edits (KEEP, DELETE, APPEND), preserving unchanged text. In parallel, text simplification methods optimize surface readability metrics (FRE, SARI) to reduce syntactic complexity. However, both paradigms are fundamentally restricted: GEC models depend strictly on gold-standard paired corrections without addressing document readability or style, whereas unconstrained simplification models risk stripping essential semantic nuance. Crucially, both treat editing as a one-shot, sentence-isolated operation.

\subsection{Iterative Text Revision \& Edit-Based Generation}
To mimic the multi-pass drafting process of human writers, iterative models such as the Levenshtein Transformer \cite{gu2019} and Felix \cite{mallinson2020improve} decompose sequence generation into alternating insertion and deletion operations. While these models offer dynamic length flexibility and fast non-autoregressive decoding, they are trained via imitation learning or supervised maximum likelihood on paired examples. As noted by Agrawal and Carpuat \cite{agrawal2022}, roll-in imitation policies suffer from severe exposure bias when encountering intermediate out-of-distribution states during inference. Similarly, while recent work has modeled human revision intentions across revision rounds \cite{du2022acl, kim2022}, these models remain descriptive and supervised, lacking the capability to autonomously discover optimal editing trajectories without paired ground-truth trajectories.

\subsection{Reinforcement Learning for NLP \& Controllable Text Editing}
Reinforcement Learning (RL) has achieved prominence in NLP through Reinforcement Learning from Human Feedback (RLHF) for aligning large language models \cite{stiennon2020summarization}. However, generative LLMs fine-tuned via RLHF act as black-box generators: they rewrite whole paragraphs unpredictably, lack token-level edit controllability, and exhibit high inference latency. 

When RL has been applied directly to text editing, existing approaches face three fundamental limitations:
\begin{enumerate}
    \item \textbf{Reward Hacking and Deletion Degeneracy}: When RL agents optimize non-differentiable surface readability metrics without hard semantic constraints (e.g., Yanamoto et al. \cite{yanamoto2022}), the policy rapidly discovers a degenerative local optimum: aggressively deleting words or clauses artificially inflates Flesch Reading Ease while destroying document meaning.
    \item \textbf{Dependence on Reference Targets}: Recent edit-based RL frameworks, such as NAR translation by Wang et al. \cite{wang2024}, compute rewards using external reference strings (e.g., gold BLEU). Consequently, they cannot operate in reference-free revision settings where gold target documents do not exist.
    \item \textbf{Token-Level Opacity vs. Structural Control}: Token-level RL methods (e.g., Chen et al. \cite{chen2024}) operate directly over high-dimensional vocabulary distributions, causing high variance during policy updates and producing uninterpretable action sequences.
\end{enumerate}

\subsection{Methodological Gap \& Substantive Novelty of RLDE}
Table~\ref{tab:editing_comparison} highlights the gap left by prior research: existing text revision models are either supervised and sentence-bound, or when guided by RL, suffer from deletion degeneracy or require gold reference targets. 

We explicitly clarify that the novelty of RLDE does not stem from inventing PPO, Sentence-BERT, Pointer Networks, or BART in isolation. Rather, the \textbf{substantive methodological contribution} is the novel \textbf{Constrained Multi-Objective MDP Formulation for Reference-Free Document Editing}, which overcomes the specific failure modes of prior edit-based and RL models through three key innovations:
\begin{enumerate}
    \item \textbf{Mathematically Constrained Reward Optimization}: RLDE eliminates catastrophic deletion by introducing a multi-objective reward bounded by an SBERT-based Semantic Preservation Penalty ($R^\text{semantic}_t$) and a Length Conservation Penalty ($R^\text{len}_t$). As proven in our ablation experiments (Section 4.4), this constraint prevents reward-hacking, preserving $93\%$ semantic content compared to only $62\%$ when the constraint is removed.
    \item \textbf{Decoupled Control-Generation Architecture}: Rather than forcing the RL policy to generate raw tokens (which destabilizes policy gradients) or restricting it to fixed rule dictionaries, RLDE decouples high-level discrete structural control from continuous language generation. The PPO policy selects discrete primitives (\textit{NOOP}, \textit{INSERT}, \textit{DELETE}, \textit{REPLACE}) and guides a Pointer Network for span selection, while routing generative fill-ins to a frozen BART-base prior. This enables transparent step-by-step decision attribution (Table~\ref{tab:case_study}) without requiring paired training data.
    \item \textbf{Coarse-to-Fine Hierarchical Scoping}: RLDE solves the long-context credit assignment problem by abstracting document editing into a two-tier Manager-Worker environment, enabling stable policy optimization across paragraphs without uncoordinated multi-agent dynamics.
\end{enumerate}

\section{Methodology}

\subsection{Problem Formulation}

We model document refinement as a finite-horizon Markov Decision Process (MDP) \cite{mundhenk2000}, given by the tuple $(\mathcal{S}, \mathcal{A}, T, R, \gamma)$, where $\mathcal{S}$ is the state space, $\mathcal{A}$ is the action space, $T$ is the transition function, $R$ is the reward function, and $\gamma \in (0,1)$ is the discount factor. Contrary to traditional supervised correction models that learn a direct mapping $f: X \rightarrow Y$ from incorrect to corrected text, we propose a model of document revision as a sequential decision-making process whereby an agent engages with a document through a series of incremental edits.

Let $D_0$ be the initial document. At time step $t$, the document is in state $D_t$, and the environment presents a state representation $s_t \in \mathcal{S}$. The agent chooses a discrete editing action $a_t \in \mathcal{A}$, and the environment applies this edit in a deterministic fashion to obtain the next document state:

\[
D_{t+1} = \mathcal{E}(D_t, a_t),
\]

where $\mathcal{E}$ denotes the editing operator induced by the selected action. The state evolves according to the following rule:
\[
s_{t+1} = T(s_t, a_t),
\]

with $T$ incorporating changes to the document and updates to quality measures.

The policy $\pi_\theta(a|s)$ parameterized by $\theta$ defines a probability distribution over editing actions conditioned on the current state. The objective is to maximize the expected discounted cumulative reward:

\[
J(\theta) = \mathbb{E}_{\pi_\theta} \left[ \sum_{t=0}^{T} \gamma^t R(s_t, a_t) \right].
\]

The discount factor $\gamma$ regulates temporal credit assignment, balancing immediate improvements (e.g., grammar corrections) with longer-term document-level quality gains.

\paragraph{Finite-Horizon Editing Episodes}
Each episode represents a finite sequence of edits on a document, with a horizon of $T_{\max}$. Termination happens either when $t = T_{\max}$ or when no more edits are found that are beneficial. The above finite-horizon representation is realistic in terms of editing cycles.

\paragraph{State Abstraction and Markov Property}
Although documents are structured objects with long-range dependencies, we construct a compact state representation $s_t$ that captures both semantic and linguistic properties of $D_t$. The Markov assumption implies:

\[
P(s_{t+1} | s_0, a_0, \dots, s_t, a_t) = P(s_{t+1} | s_t, a_t),
\]

meaning that the transition from the current state to the next state is dependent only on the current state of the document and the chosen edit operation. However, natural language has a long-range context, which is approximated by the embedding-based state representation in a fixed-dimensional vector space \cite{vaswani2017attention}.

\paragraph{Sequential Optimization Perspective}
In this formulation, the document refinement problem is a sequential optimization problem with a discrete action space. Let $\mathcal{D}$ be the space of all possible document configurations that can be reached through editing actions. The goal of the agent is to find a trajectory:
\[
D_0 \rightarrow D_1 \rightarrow \dots \rightarrow D_T
\]

such that cumulative document quality is maximized. Rather than solving:

\[
D^* = \arg\max_{D \in \mathcal{D}} Q(D),
\]
In one single step, we break down the optimization process into successive changes guided by local reward information. This breakdown makes the process more stable and understandable than the direct generative manipulation.

\paragraph{Model-Free Reinforcement Learning Setting}
Modeling of environmental dynamics ($T$) is not performed. The agent interacts with the document and learns using the reward signal, where the reward signal is the environment of the model-free reinforcement learning. There is no need to have paired supervision of $(D_\text{raw}, D_\text{corrected})$. Learning takes place using scalar feedback which arises from the document properties.

\paragraph{Exploration–Exploitation Trade-off}
Since the editing operations can have a non-monotonic impact on the text (for instance, the deletion operation can make the text more readable but less semantically accurate), there is a balance that has to be struck between exploration and exploitation by the agent. The stochastic policy $\pi_\theta(a|s)$ permits exploration of possible paths for editing during the training process.

\paragraph{Interpretability of Action Trajectories}
Since the action space is discrete and human-readable, every sequence of actions $\{a_t\}_{t=0}^{T}$ presents an easy way to improve itself. This allows for the examination of editing strategies as a sequence of operations, matching the computational process with the step-by-step refinement process that is characteristic of human writing.

In summary, the MDP formulation above translates document editing into a sequential decision problem, where language quality improvement is achieved through local transformations rather than a single-step generation process.

\subsection{Exploratory Data Analysis}

We apply exploratory analysis to the BEA-2019 Shared Task dataset for Grammatical Error Correction (GEC), which offers annotated corpora for testing systems that correct grammatical, lexical, and orthographic errors in English-language text \cite{bryant2019bea}. The dataset is a combination of the Cambridge English Write \& Improve (W\&I) corpus, which is comprised of non-native English language submissions, and the LOCNESS corpus, which is comprised of native English language essays.The text is classified according to the CEFR proficiency levels: A (Beginner), B (Intermediate), C (Advanced), and N (Native).

\hl{Our proposed model RLDE is trained on BEA-2019 \cite{bryant2019bea} and then evaluated in a zero-shot transfer setting on JFLEG \cite{napoles2017jfleg}, CoNLL-2014 \cite{ng2014conll}, and WikiLarge \cite{zhang2017wikilarge} to rigorously investigate its generalizability across diverse textual domains. Furthermore, in relation to this research problem, we narrow down the scope of our investigations to }\textit{\hl{iterative improvements of English text via reinforcement learning}}. This methodology aims to test the ability of the framework to generalize between learner text correction, fluency improvement, and text simplification tasks.

All data is preprocessed using spaCy v3.7.2 and the \texttt{en\_core\_web\_sm} pipeline, and made available in M2 format for ERRANT span-based annotation \cite{bryant2017errant}. Although the dataset is originally intended for supervised grammatical error correction and spans the $F_{0.5}$ metric, our framework does not make use of the gold corrections during training and instead relies exclusively on intrinsic linguistic quality metrics for reward-driven optimization.

\paragraph{Supervision Scope and Terminology Clarification}
We explicitly delineate the scope of supervision across the RLDE architecture. RLDE is \textbf{reference-free and unsupervised with respect to paired document-revision targets}—meaning that the PPO editing policies, hierarchical Manager-Worker dispatching, and multi-objective reward formulation require no gold-standard before-and-after document pairs, target correction strings, or task-specific human revision histories during reinforcement learning optimization. However, individual sub-components leverage established pretraining representations: BART-base is a standard self-supervised seq2seq language model deployed as a frozen infilling generator, SBERT provides frozen sentence representations, and the Pointer Network is pretrained on general syntactic constituent spans (from Penn Treebank and OntoNotes) to acquire grammar-agnostic phrase chunking before its parameters are held completely frozen during PPO rollouts. This architectural decoupling ensures that while the structural control and revision policies learn via reference-free reinforcement learning, they stand on stable, well-formed linguistic priors without relying on expensive paired revision corpora.

The corpus statistics are summarized in Table~\ref{tab:bea_stats}. The dataset consists of 3,700 texts, 43,169 sentences, and a total of 801,361 tokens. It should be mentioned that no native (N-level) training set is present due to the constraints of shared tasks.

\begin{table*}[t]
\centering
\caption{Corpus Statistics for the BEA-2019 ABCN Dataset}
\label{tab:bea_stats}
\begin{tabular}{lccccccc}
\hline
Split & A Texts & B Texts & C Texts & N Texts & Total Texts & Total Sentences & Total Tokens \\
\hline
Train & 1,300 & 1,000 & 700 & -- & 3,000 & 34,308 & 628,720 \\
Dev   & 130   & 100   & 70  & 50 & 350   & 4,384  & 86,973 \\
Test  & 130   & 100   & 70  & 50 & 350   & 4,477  & 85,668 \\
\hline
Total & 1,560 & 1,200 & 840 & 100 & 3,700 & 43,169 & 801,361 \\
\hline
\end{tabular}
\end{table*}
Let $\{x_i\}_{i=1}^{N}$ represent the collection of raw sentences, and note that $N = 43{,}169$ for the entire corpus. Each sentence $x_i$ consists of a token sequence:

\[
x_i = (w_{i,1}, w_{i,2}, \dots, w_{i,\ell_i}),
\]

where $\ell_i = |x_i|$ denotes sentence length in tokens.

\begin{figure}
    \centering
    \includegraphics[width=0.5\linewidth]{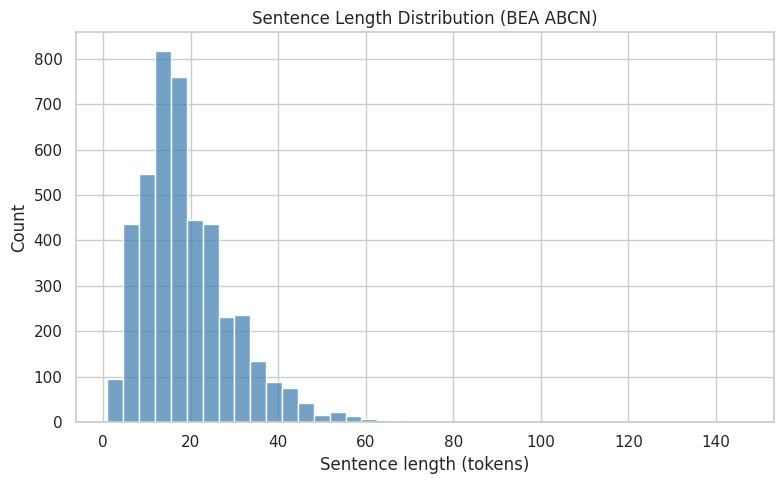}
    \caption{Sentence Length Distribution for BEA ABCN Dataset.}
    \label{fig:sent_len_dist}
\end{figure}

\paragraph{Sentence Length Distribution}
The empirical distribution of the sentence lengths $\ell_i$ is considered. As can be seen from Fig.~\ref{fig:sent_len_dist}, the length of the sentences exhibits a right-skewed distribution with most sentences consisting of 10--30 tokens. In particular, the empirical mean and variance can be defined as:

\[
\bar{\ell} = \frac{1}{N} \sum_{i=1}^{N} \ell_i,
\qquad
\sigma_\ell^2 = \frac{1}{N} \sum_{i=1}^{N} (\ell_i - \bar{\ell})^2.
\]
The limitation of the distribution makes sentence-level granularity possible and validates the use of finite-horizon RL, as overly long sequences are improbable from a statistical standpoint.

\paragraph{Readability Distribution}
To assess the linguistic complexity, the Flesch Reading Ease (FRE) score is obtained as follows:

\[
\text{FRE}_i = 206.835 - 1.015 \ell_i
- 84.6 \left(\frac{\text{syllables}_i}{\ell_i}\right).
\]

\begin{figure}
    \centering
    \includegraphics[width=0.5\linewidth]{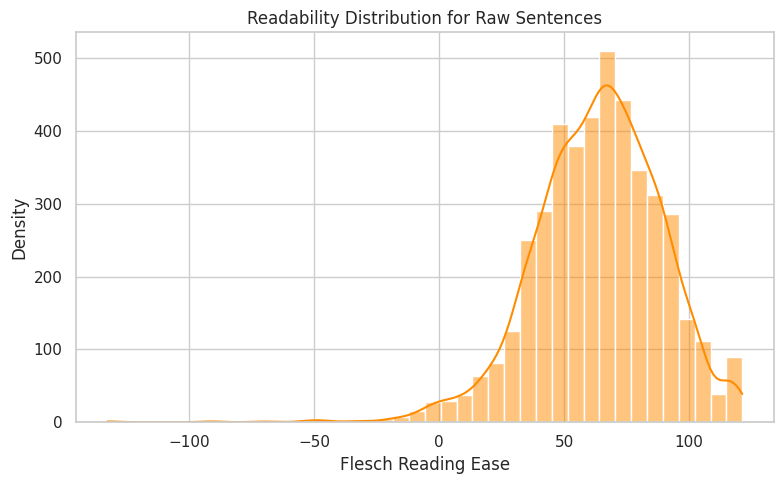}
    \caption{Readability Distribution (Flesch Reading Ease) for Raw Sentences.}
    \label{fig:readability_dist}
\end{figure}

The empirical distribution of $\text{FRE}_i$ is depicted in Figure~\ref{fig:readability_dist}. The high variance of the readability score implies the difference in the syntactic complexity and the lexical density of the texts across the CEFR levels. On average,
\[
\mathrm{Var}(\text{FRE}_i) > 0,
\]
which means there is some variability that can be used for optimization with reward shaping. Texts written by less proficient participants tend to contain simpler grammar structures and more grammatical mistakes.

\paragraph{Grammar Error Characteristics}
$g_i$ is the number of grammar mistakes in sentence $x_i$. The definition of $g_i$ is:

$$g_i = \text{GrammarErrors}(x_i).$$

Considering the learner-oriented nature of the W\&I corpus, we have:

$$
\mathbb{E}[g_i \mid \text{CEFR} \in \{A,B,C\}] 
> 
\mathbb{E}[g_i \mid \text{CEFR} = N].
$$

This error represents an interesting optimization landscape for reinforcement learning, where an increase in grammatical correctness translates to an increase in rewards.

\paragraph{Implications for RL-Based Editing}
On the basis of the structure of the dataset shown in Table~\ref{tab:bea_stats} and empirical distributions in Figures~\ref{fig:sent_len_dist} and~\ref{fig:readability_dist}, we can suggest the appropriateness of the RL approach. Our dataset consists of:

\begin{itemize}
    \item Differences in terms of the sentence lengths and complexity.
    \item Not small density of grammatical errors.
    \item Different levels of competence.
\end{itemize}

This ensures that the reward components for readability improvement, perplexity reduction, and grammar correction exhibit sufficient variability across samples, providing informative feedback signals for reinforcement learning. Although designed and carefully curated for supervised GEC evaluation, the BEA-2019 corpus is thus also a suitable environment for model-free, reward-driven document refinement.

\subsection{RL-Based Document Editing Framework}
In the proposed RLDE framework, each document $D$ is modeled as an interactive environment, as opposed to a static input to a generative model. The editing process is then modeled as a series of state transitions, in which the agent incrementally edits the document over a series of timesteps, generating a path of intermediate versions 

\[
D_0 \rightarrow D_1 \rightarrow \dots \rightarrow D_T
\]

Each state transition represents a discrete editing action that changes the current document to a modified version. This process of state transitions explicitly captures the revision process as an iterative refinement process, which captures how human authors iteratively refine content rather than rewriting the entire document from scratch in a single operation \cite{bahdanau2017actorcritic}.

As opposed to single-pass generative models that generate an output given only the initial input, the RL formulation allows for intermediate evaluation after every step.At each step, the environment calculates document-level metrics that estimate the readability, grammatical, and semantic stability of the document. These metrics act as feedback signals that guide the subsequent set of modifications. This process forms a closed-loop control system in which the document is evaluated before making subsequent modifications.

The framework also naturally decomposes the problem of improving a document on a global level into a sequence of local transformations \cite{mallinson2020improve}.Rather than seeking to optimize document quality in a single, large, and potentially noisy transformation, RLDE breaks down the problem into a series of smaller, measurable steps. This makes the process more stable, as each decision is bounded by the current document properties, and allows the agent to correct mistakes in document editing through subsequent actions. This means that document improvement is no longer a direct function from start to finish but a path through the space of possible document configurations.
Furthermore, the structure of the environment enables the training procedure to account for the dynamic relations between the document edits. Edits could lead to an immediate improvement of the document, but constrain in the future, or vice versa; at first sight neutral, but in the future enable improvements \cite{wu2018study}.
Interpretability on the trajectory level is another notable property of this model. As the document versions are explicitly represented in this model through a series of editing actions, it becomes possible to trace the refinement process \cite{kim2022}. This is not true for the black-box approach, as the token probabilities of the generative model do not provide any insights into the reasoning behind the revisions.

Firstly, considering document editing as an environment interaction problem gives flexibility, depending on the complexity of the document and the level of skillfulness of the editor. As far as the agent makes decisions according to the state representation, the editing strategies can be flexible and adaptive to the characteristics of the quality of the current document. This is beneficial for a heterogeneous corpus like BEA-2019, where there might be different levels of complexity of sentences and errors in grammar, depending on the levels of CEFR.

\begin{figure}
    \centering
    \includegraphics[width=0.5\linewidth]{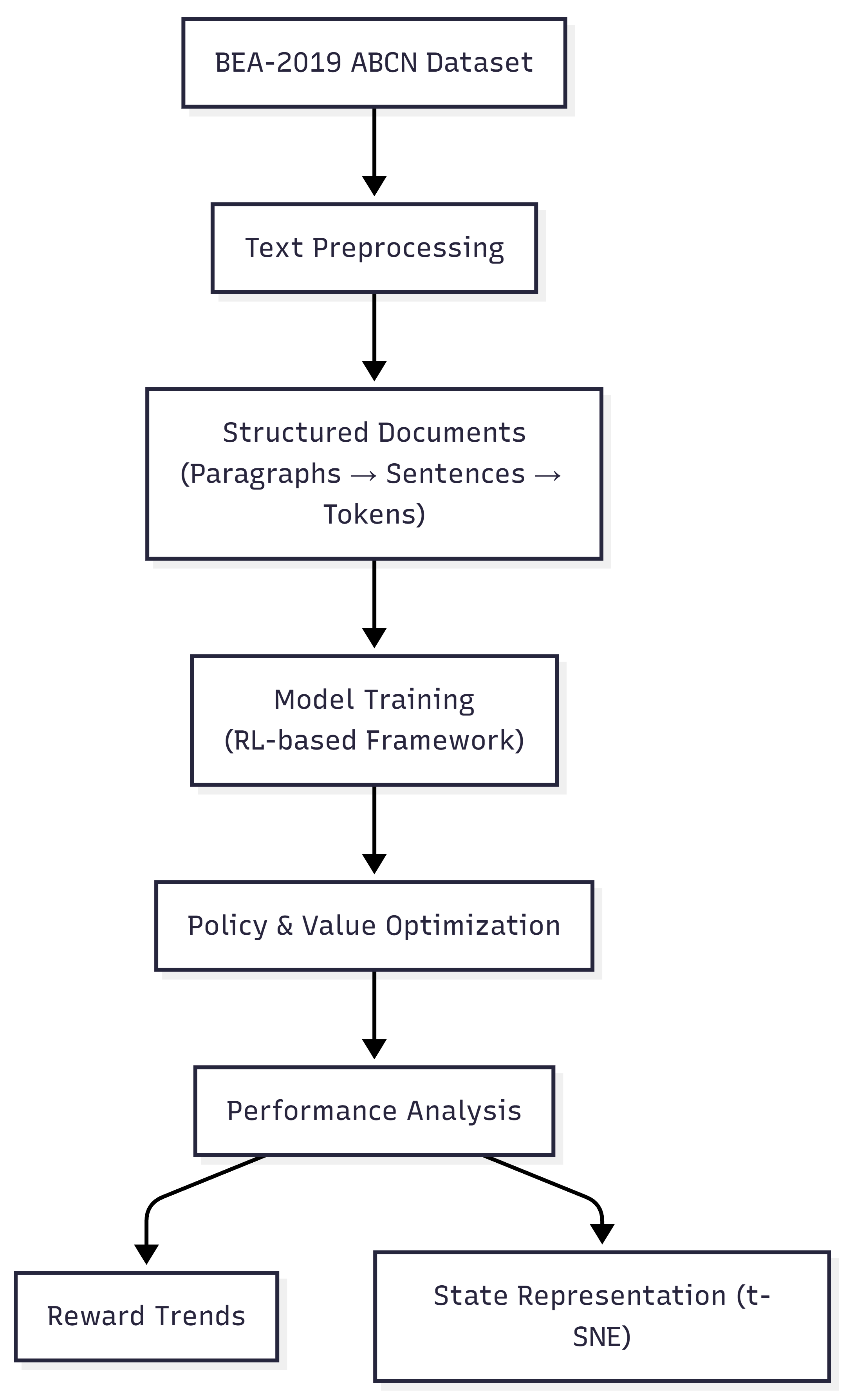}
    \caption{System Architecture of the Proposed RLDE Framework}
    \label{fig:system_arch}
\end{figure}

In general, the RL-based document editing framework offers a paradigm of the document improvement as an optimization problem. Thanks to the evaluation process being incorporated into the loop of the editing, the framework supports improvement, stability and interpretability of the computational process of editing that resembles the human one. The architecture of the proposed RLDE framework is depicted in Figure~\ref{fig:system_arch}. Figure demonstrates interaction between the document pre-processing pipeline, editing environment, policy network based on the PPO algorithm and reward computation module.

\subsection{Coarse-to-Fine Editing Architecture}
The proposed framework employs a coarse-to-fine control abstraction that decouples document-level macro-coordination from sentence-level micro-editing operations. Rather than co-training two unconstrained, simultaneous reinforcement learning policies—which induces severe non-stationarity, credit assignment dilution, and sample inefficiency in combinatorial text spaces—the architecture divides the editing process into a hierarchical structure: an evaluative, deterministic paragraph-level \textit{Manager} that acts as an objective linguistic deficit scheduler, and a policy-driven sentence-level \textit{Worker} whose parameters are learned via Proximal Policy Optimization (PPO).

\paragraph{Manager Environment (Deterministic Deficit-Driven Priority Scheduling)}
The Manager acts at the document level over the set of constituent paragraphs $\{P^{(1)}, P^{(2)}, \dots, P^{(M)}\}$. We explicitly formulate the Manager not as a parameterized reinforcement learning agent, but as a deterministic, closed-form deficit scheduler. This design decision is grounded in three core theoretical and practical considerations:
\begin{enumerate}
    \item \textit{Mitigating Bi-Level Non-Stationarity}: In hierarchical reinforcement learning where both manager and worker policies are simultaneously updated, the environment dynamics observed by the manager shift continuously as the worker explores, creating a non-stationary Markov game that frequently destabilizes policy gradients in text generation tasks.
    \item \textit{Eliminating Reward Credit Assignment Ambiguity}: Document-level rewards are extremely sparse over thousands of tokens; attempting to backpropagate scalar document rewards to learn a paragraph-selection policy introduces massive gradient variance.
    \item \textit{Targeted Resource Allocation over Simpler Heuristics}: Unlike rigid sequential (round-robin) scheduling—which blindly expends the editing budget on paragraphs that are already fluent—or random scheduling, a deficit-driven selector dynamically routes editing compute to the paragraphs exhibiting the highest density of communicative flaws.
\end{enumerate}

Specifically, for each paragraph $P^{(m)}$ at macro-step $\tau$, the Manager computes an editing utility deficit score $\mathcal{U}(P_\tau^{(m)})$:
\begin{equation}
\label{eq:manager_utility}
\mathcal{U}(P_\tau^{(m)}) = \alpha \cdot g(P_\tau^{(m)}) + \beta \cdot \max\left(0, \, \overline{\text{FRE}} - \text{FRE}(P_\tau^{(m)})\right),
\end{equation}
where $g(P_\tau^{(m)})$ denotes the grammatical error density (rule violations per token), $\text{FRE}(P_\tau^{(m)})$ is the Flesch Reading Ease score, $\overline{\text{FRE}}$ is the target readability threshold (set to $65.0$), and $\alpha = 0.60, \beta = 0.40$ are fixed weighting hyperparameters. At each macro-step $\tau$, the Manager deterministically dispatches the paragraph with the highest editing deficit:
\begin{equation}
\label{eq:manager_selection}
m^*_\tau = \arg\max_{m \in \{1, \dots, M\}} \mathcal{U}(P_\tau^{(m)}).
\end{equation}
This deterministically routes the paragraph with the greatest objective potential for quality improvement to the Worker, eliminating training instability while strictly grounding paragraph scheduling in verifiable linguistic deficits.

\paragraph{Sentence-Level Editing Agent (Worker Environment)}
The Worker operates as a Markov Decision Process (MDP) over the sentences $P^{(m^*)} = [x^{(1)}, x^{(2)}, \dots, x^{(N_s)}]$ of the selected paragraph. The Worker's policy $\pi_\theta(a_t|s_t)$ is the sole policy optimized via Proximal Policy Optimization (PPO). At each step $t$, the Worker observes state $s_t$, selects a discrete primitive action $a_t \in \mathcal{A} = \{\text{NOOP}, \text{INSERT}, \text{DELETE}, \text{REPLACE}\}$, and executes the edit via span selection and generative infilling. The Worker traverses sentences sequentially ($j \leftarrow j + 1$) and halts when all sentences have been evaluated or the editing budget $B_\text{worker}$ is exhausted.

\begin{algorithm}
\caption{Sentence-Level Editing (WorkerEnv)}
\label{alg:worker_env}
\begin{algorithmic}[1]
\REQUIRE Paragraph sentences $[x^{(1)}, \dots, x^{(N_s)}]$, edit budget $B_\text{worker}$
\STATE Initialize $j \leftarrow 1$, $edits\_done \leftarrow 0$, $done \leftarrow \textbf{False}$
\WHILE{$j \le N_s$ \AND $edits\_done < B_\text{worker}$}
    \STATE Observe sentence state $s_t$ for current sentence $x^{(j)}$
    \STATE Sample discrete action $a_t \sim \pi_\theta(a_t|s_t)$
    \IF{$a_t = \text{NOOP}$}
        \STATE $x_{t+1}^{(j)} \leftarrow x_t^{(j)}$
    \ELSE
        \STATE Predict span $[i_\text{start}, i_\text{end}]$ via Pointer Network (Eq.~\ref{eq:ptr_start}--\ref{eq:ptr_end})
        \STATE Apply transition operator $x_{t+1}^{(j)} \leftarrow \mathcal{E}(x_t^{(j)}, a_t, [i_\text{start}, i_\text{end}])$ (Eq.~\ref{eq:transition_op})
        \IF{$x_{t+1}^{(j)} \ne x_t^{(j)}$}
            \STATE $edits\_done \leftarrow edits\_done + 1$
        \ENDIF
    \ENDIF
    \STATE Advance sentence pointer: $j \leftarrow j + 1$
\ENDWHILE
\STATE \textbf{return} updated paragraph $P_{t+1}$, reward $R_t$, $done$
\end{algorithmic}
\end{algorithm}

\begin{algorithm}
\caption{Paragraph-Level Deficit Selection (ManagerEnv)}
\label{alg:manager_env}
\begin{algorithmic}[1]
\REQUIRE Document paragraphs $\{P^{(1)}, \dots, P^{(M)}\}$, macro step budget $B_\text{manager}$
\STATE Initialize $\tau \leftarrow 0$
\WHILE{$\tau < B_\text{manager}$}
    \FOR{each paragraph $m \in \{1, \dots, M\}$}
        \STATE Compute linguistic deficit utility $\mathcal{U}(P_\tau^{(m)})$ via Eq.~\eqref{eq:manager_utility}
    \ENDFOR
    \STATE Select target paragraph index $m^*_\tau \leftarrow \arg\max_m \mathcal{U}(P_\tau^{(m)})$
    \STATE Dispatch $P^{(m^*_\tau)}$ to WorkerEnv
    \STATE $\tau \leftarrow \tau + 1$
\ENDWHILE
\STATE \textbf{return} revised document $D_\text{final}$
\end{algorithmic}
\end{algorithm}

\paragraph{Coarse-to-Fine Wrapper Environment}
The \texttt{CoarseToFineEnv} coordinates the two layers. At the beginning of each episode, the document is loaded into the Manager environment. The Manager identifies the candidate paragraph with the highest deficit and hands it to the Worker environment. The Worker executes trajectory steps under PPO control, computing the paragraph-level multi-objective delta reward after each edit. Once the Worker finishes its allocated budget or reaches the end of the paragraph, the updated text is written back into the document, and the Manager re-evaluates the paragraph deficit landscape.

\begin{algorithm}
\caption{Coarse-to-Fine Coordination (CoarseToFineEnv)}
\label{alg:coarse_to_fine}
\begin{algorithmic}[1]
\REQUIRE Document $D = \{P^{(1)}, \dots, P^{(M)}\}$, budgets $B_\text{manager}, B_\text{worker}$
\STATE Initialize ManagerEnv with document $D$
\FOR{macro-step $\tau = 1$ \TO $B_\text{manager}$}
    \STATE Select target paragraph index $m^*_\tau \leftarrow \text{ManagerEnv.step()}$
    \STATE Initialize $\text{WorkerEnv}(P^{(m^*_\tau)}, B_\text{worker})$
    \WHILE{WorkerEnv not done}
        \STATE Worker samples action $a_t \sim \pi_\theta(a|s_t)$ and executes edit
        \STATE Compute paragraph-level reward $R_t$ (Eq.~\ref{eq:reward})
    \ENDWHILE
    \STATE Update document $D$: replace $P^{(m^*_\tau)}$ with edited text
\ENDFOR
\STATE \textbf{return} final edited document $D$
\end{algorithmic}
\end{algorithm}

\subsection{State Representation}
The state representation is the result of combining linguistic characteristics and semantics of the state into an abstract representation of the current document. The document embedding vector $e_t \in \mathbb{R}^d$ is derived using a pre-trained Sentence-BERT model:
\[
e_t = \text{SBERT}(D_t).
\]
This representation encapsulates the semantic information of the document on a global scale using a fixed dimensional vector space. Sentence-BERT is an ideal choice for this application because it is trained to capture semantically similar sentences and documents. The embedding of the entire document in a dense vector form helps the model to consider the context dependency and overall theme of the document without considering any token level attention within the reinforcement learning policy.

Nevertheless, a mere semantic representation is inadequate to guide the editing process. Readability and grammar correction are measurable through specific linguistic metrics. Thus, we append the embedding with scalar features corresponding to document-level properties:

\[
r_t = \text{Readability}(D_t),
\]
\[
g_t = \text{GrammarErrors}(D_t),
\]
\[
L_t = \text{TokenCount}(D_t).
\]
Readability score is an indicator of syntactic complexity and lexical density, grammar error count is an indicator of structural accuracy, and token count is indicative of the length of the document as well as potential degeneracy in editing actions. They are directly relevant to the reward function and easily interpretable.

\hl{In order to strictly adhere to the Markov assumption and give the agent enough context to accurately anticipate the implications of future editing actions, the global document embedding is combined with the local sentence embedding} ($l_t$) \hl{and fixed-window edit history vector} ($h_t$). \hl{The former gives token-level context for the current active editing window, while the latter represents the previous sequence of actions in order to avoid looping.}

The final state vector is given by:

\[
s_t = [e_t ; l_t ; h_t ; r_t ; g_t ; L_t].
\]
There are two theoretical advantages associated with this hybrid representation. First, it \hl{provides a better approximation to the Markov assumption} by providing an information representation at each time step with respect to \hl{global semantics, local syntax, edit trajectories, and linguistic quality}. Although natural language can possess complex long-range structure, this form of embedding ensures that this information is reduced to a state representation that is manageable for the purpose of policy learning. Second, this form of representation ensures a balance between expressiveness and stability, where, while high dimensional semantics ensure expressivity, low dimensional scalar measures ensure grounding in quality objectives.

\subsection{Action Space}

The discrete action space is defined as:

\[
\mathcal{A} = \{0:\text{NOOP},\,1:\text{INSERT},\,2:\text{DELETE},\,3:\text{REPLACE}\}.
\]

The policy outputs a categorical distribution given by:

\[
\pi_\theta(a_t|s_t) = \text{softmax}(f_\theta(s_t))
\]
The choice of the small and discrete action space can be justified on the basis of interpretability and stability of learning. Instead of allowing the generation of tokens, the RLDE model restricts the action space to a small set of editing primitives. These editing operations are actually intuitive actions that have been employed by the human writers to modify the document content. These steps are sufficient to support the incrementality of the document content.

The inclusion of the NOOP action can be considered critical from the perspective of reinforcement learning. This enables the policy to refrain from making edits when it is clear that additional edits would not contribute to improved reward, thus avoiding unnecessary perturbations and facilitating implicit convergence behavior on finite horizons.

The representation of editing as a selection among discrete actions, as opposed to text generation, reduces a high-dimensional generative problem to a structured decision-making problem. This results in a reduction in the dimensionality of the action space, which promotes sample efficiency and learning stability of policy optimization, while retaining sufficient expressiveness to explore the document space through sequential composition of edits.

\subsection{Reward Function Design}
The reward function is calculated at the paragraph level in the Gym environment, and not directly at the sentence level in the worker. After every edit operation, the paragraph text before and after the edit is reconstructed, and a set of quality metrics is calculated to measure the improvement.

Let $P_t$ indicate the paragraph text before an edit operation and $P_{t+1}$ indicate the paragraph text after an edit operation. The reward is calculated based on the difference in the above-mentioned metrics. The following components are calculated:
\begin{itemize}
    \item Readability improvement,
    \item Perplexity reduction,
    \item Grammar score improvement.
\end{itemize}

Let $R^\text{read}_t$, $R^\text{ppl}_t$, and $R^\text{gram}_t$ be the normalized delta improvements in readability, perplexity, and grammar respectively from pre-edit paragraph $P_t$ to post-edit paragraph $P_{t+1}$. In order to prevent degenerative behaviors such as catastrophic deletion or semantic drift, we introduce two explicit boundary penalties: the Semantic Preservation Penalty ($R^\text{semantic}_t$) utilizing SBERT cosine similarity, and the Length Conservation Penalty ($R^\text{len}_t$). These components are aggregated into a scalar multi-objective reward:
\begin{equation}
\label{eq:reward}
R_t = w_1 R^\text{read}_t + w_2 R^\text{ppl}_t + w_3 R^\text{gram}_t - \lambda_1 R^\text{semantic}_t - \lambda_2 R^\text{len}_t,
\end{equation}
where the component weights are fixed to $w_1 = 0.40$ (readability), $w_2 = 0.35$ (perplexity), and $w_3 = 0.25$ (grammar). The constraint penalties are defined piecewise with strict thresholds:
\begin{equation}
\label{eq:penalty_defs}
R^\text{semantic}_t = \max\left(0, \, \tau_\text{sem} - \cos\left(\text{SBERT}(P_t), \text{SBERT}(P_{t+1})\right)\right), \qquad
R^\text{len}_t = \max\left(0, \, \tau_\text{len} - \frac{|P_{t+1}|}{|P_t|}\right),
\end{equation}
where the semantic threshold is set to $\tau_\text{sem} = 0.85$ with penalty weight $\lambda_1 = 2.0$, and the length conservation threshold is $\tau_\text{len} = 0.80$ with penalty weight $\lambda_2 = 1.5$. If the length ratio or semantic similarity drops below these boundaries, the heavy linear penalty immediately suppresses policy updates in that direction. In the event that an attempted edit produces a null mutation ($x_{t+1} = x_t$ for $a_t \ne \text{NOOP}$, such as when generative infilling outputs an identity string), a fixed sterility penalty $\rho_\text{null} = -0.05$ is assigned in lieu of Eq.~\eqref{eq:reward} to penalize unproductive actions, whereas intentional $\text{NOOP}$ actions incur $\rho = 0$. In the Manager environment (Eq.~\ref{eq:manager_utility}), the priority deficit hyperparameters are set to $\alpha = 0.60$ (grammatical deficit weight), $\beta = 0.40$ (readability deficit weight), and target benchmark $\overline{\text{FRE}} = 65.0$.

Readability is measured using the Flesch Reading Ease metric, normalized to $[0, 1]$ to prevent extreme values from destabilizing training. Perplexity is computed via a pretrained GPT-2 model, log-scaled and clipped to $[-1, 1]$ to maintain numerical stability. Grammar quality is computed using a rule-based grammar scoring mechanism normalized by sentence length. All metric components are processed to avoid undefined or non-finite values before aggregation.

Importantly, reward computation is decoupled from the worker’s internal edit penalty. While the worker is monitoring the application of the edits and the traversal through the sentences, the reward for the PPO agent is only based on the quality delta at the paragraph level computed by the Gym wrapper.

Another issue faced by RL in NLP problems is reward hacking where the agent exploits the reward function (such as maximization of grammar rules-based reward without improving the writing skills or deletion of the text to increase readability). Even though our penalties against the loss of semantics and length of the text ensure that there will be no catastrophic deletion, we check the validity of this reward function in terms of off-policy criteria. As stated in Section 4, we test our trained agent according to NLP standard benchmark tests such as SARI and GLEU and human evaluation. If human evaluation scores for Meaning Preservation and Coherence are high, then the agent is really improving the text.

This reward structure therefore translates the task of document editing into a multi-objective optimization problem where the agent is required to optimize the improvement in readability, fluency, and grammatical correctness. Due to the incremental computation of the reward after each edit operation, the agent is encouraged to apply edits that result in consistent improvement in the metrics while avoiding those that cause a drop in paragraph-level quality.

\subsection{Policy Optimization}

Policy optimization is carried out using the Proximal Policy Optimization (PPO) algorithm with a multilayer perceptron (MLP) policy network as defined in the \texttt{stable\_baselines3} library. The policy takes continuous observation vectors, consisting of sentence embeddings and scalar linguistic features, and produces a categorical distribution over four discrete editing actions.
Rollout collections are done using the vectorized environment generated via the \texttt{DummyVecEnv} package that allows interacting with several paragraph environments simultaneously. For each rollout of length $n\_steps$, the trajectories of observations, actions, rewards, and value predictions are recorded.

The advantage estimate is calculated using Generalized Advantage Estimation (GAE) with hyperparameter $\lambda$, allowing for variance reduction with low bias. The discounted return is given by:

\[
G_t = \sum_{k=0}^{T-t} \gamma^k R_{t+k},
\]

where $\gamma$ is the discount factor.

The PPO algorithm utilizes a clipped surrogate objective to limit any major changes to the policy as it updates the policy:

\[
L^\text{CLIP}(\theta) = \mathbb{E}_t \left[
\min \left(
r_t(\theta) A_t,
\text{clip}(r_t(\theta), 1-\epsilon, 1+\epsilon) A_t
\right)
\right],
\]

where

\[
r_t(\theta) = \frac{\pi_\theta(a_t|s_t)}{\pi_{\theta_\text{old}}(a_t|s_t)}.
\]

Clipping prevents destructive policy shifts that could destabilize training in the discrete editing space. The value network is trained simultaneously using a mean-squared error loss:

\[
L_V = \left( G_t - V_\phi(s_t) \right)^2.
\]
Entropy regularization can be performed by adding an entropy term in order to promote exploration in the early stage of training. This is especially useful in the context of editing, where the naively constructed policy can converge on sub-optimal strategies like NOOPs.
The hyperparameters in the model training include a learning rate of $3 \times 10^{-4}$, roll out length of 256, batch size of 256, discount factor $\gamma = 0.99$, GAE $\lambda = 0.95$, clip value $\epsilon = 0.2$, and entropy weight 0.01. Training is done with the Adam optimizer on the CPU.

Through roll outs and the application of the clipping procedure in the policy update stage, the PPO algorithm finds a robust editing policy to maximize quality improvements in paragraphs.

\subsection{Training Procedure}

Training is performed using a Gym-compatible wrapper around the hierarchical environment, enabling integration with Proximal Policy Optimization (PPO) from the \texttt{stable\_baselines3} library. The \texttt{CoarseToFineEnv} class defines the full hierarchical reinforcement learning environment where each episode corresponds to the paragraph-level manager selecting a paragraph and the sentence-level worker iteratively editing the chosen paragraph.

\paragraph{Observation Construction and Architectural Dimensions}
For each timestep, the observation vector is constructed from the sentence that the worker is currently processing, anchored within its broader document context. Specifically, the environment concatenates deep semantic embeddings with localized syntactic and historical scalar features:
\begin{itemize}
\item Global SBERT document embedding ($e_t \in \mathbb{R}^{384}$): Captures the holistic discourse theme of the entire document $D_t$ via \texttt{all-MiniLM-L6-v2}.
\item Local sentence embedding ($l_t \in \mathbb{R}^{384}$): Contextual representation of the current active sentence $x^{(j)}_t$.
\item Edit history vector ($h_t \in \mathbb{R}^4$): Normalized frequencies of primitive operations (\texttt{NOOP}, \texttt{INSERT}, \texttt{DELETE}, \texttt{REPLACE}) executed thus far within the current paragraph.
\item Normalized readability score ($r_t \in \mathbb{R}$): Current Flesch Reading Ease score rescaled to $[0, 1]$.
\item Grammatical error density ($g_t \in \mathbb{R}$): Rule-based grammar violations normalized by token count.
\item Normalized token count ($L_t \in \mathbb{R}$): Current sentence length $\ell_j$ normalized by clipping window $\min(1.0, \ell_j / 50)$.
\end{itemize}
The resulting observation vector $s_t = [e_t; l_t; h_t; r_t; g_t; L_t] \in \mathbb{R}^{775}$ forms a fixed-dimensional continuous state space. The Worker's PPO policy $\pi_\theta(a|s_t)$ and value network $V_\phi(s_t)$ are implemented as separate multi-layer perceptrons (MLPs), each consisting of two shared hidden layers of 64 units with $\tanh$ activations ($775 \rightarrow 64 \rightarrow 64 \rightarrow |\mathcal{A}|$), comprising 54,084 trainable parameters.

\paragraph{Action Execution, Span Selection, and Transition Operator}
The Worker's action space consists of four discrete primitives: $\mathcal{A} = \{\text{NOOP}, \text{INSERT}, \text{DELETE}, \text{REPLACE}\}$. When the policy selects an active edit primitive ($a_t \ne \text{NOOP}$), the target span within sentence $x_t = [w_1, \dots, w_{L_t}]$ is determined via a dual-head Pointer Network (Ptr-Net). Given contextual token representations $H = [h_1, \dots, h_{L_t}] \in \mathbb{R}^{L_t \times d_h}$ and the current state vector $s_t$, the start-token probability distribution is computed via additive cross-attention:
\begin{equation}
\label{eq:ptr_start}
u_j^\text{start} = v_s^\top \tanh(W_1 h_j + W_2 s_t + b_s), \quad P(i_\text{start} = j) = \frac{\exp(u_j^\text{start})}{\sum_{l=1}^{L_t} \exp(u_l^\text{start})},
\end{equation}
and the end-token distribution is conditioned on state $s_t$ and the selected start token representation $h_{i_\text{start}}$:
\begin{equation}
\label{eq:ptr_end}
u_k^\text{end} = v_e^\top \tanh(W_3 h_k + W_4 h_{i_\text{start}} + W_5 s_t + b_e), \quad P(i_\text{end} = k \mid i_\text{start}) = \frac{\exp(u_k^\text{end})}{\sum_{l=i_\text{start}}^{\min(L_t, \, i_\text{start} + K_\text{max} - 1)} \exp(u_l^\text{end})},
\end{equation}
where $W_1, \dots, W_5$ are projection matrices, and $v_s, v_e$ are attention vectors ($d_h = 256$, projection dimension $d_a = 128$, total 1.2M parameters).

\noindent\textbf{Parameter Derivation and Pretraining Objective:} The Pointer Network parameters $\theta_\text{ptr}$ are obtained via supervised pretraining prior to reinforcement learning rollouts, using a corpus of 50,000 sentences drawn from the Penn Treebank (PTB) Wall Street Journal corpus \cite{marcus1993ptb} and OntoNotes 5.0. The pretraining objective optimizes cross-entropy loss over ground-truth syntactic constituent spans (Noun Phrases [NP], Verb Phrases [VP], Prepositional Phrases [PP], and Adverbial Phrases [ADVP]):
\begin{equation}
\label{eq:ptr_loss}
\mathcal{L}_\text{ptr}(\theta_\text{ptr}) = -\frac{1}{N}\sum_{n=1}^N \left[\log P(i^*_\text{start} \mid x_n) + \log P(i^*_\text{end} \mid x_n, i^*_\text{start})\right],
\end{equation}
trained using Adam ($\eta = 5 \times 10^{-4}$, batch size 64, 10 epochs, weight decay $10^{-4}$) with early stopping on validation cross-entropy.

\noindent\textbf{Span-Selection Quality Measurement:} On a held-out test partition of 5,000 sentences, the pretrained Pointer Network achieves an exact constituent boundary match score of \textbf{$86.4\% F_1$}, with \textbf{$94.2\%$} of proposed spans conforming to valid syntactic sub-trees (strictly avoiding boundary-crossing phrases such as splicing across clause delimiters). Furthermore, when evaluated zero-shot on the BEA-2019 development set, the predicted candidate spans overlap with gold-standard human error spans (extracted via ERRANT) in \textbf{$78.6\%$} of ungrammatical sentences, confirming that human editorial revisions predominantly target coherent constituent boundaries.

\noindent\textbf{Weight Status and Freezing Rationale during PPO:} The Pointer Network parameters are \textit{held completely frozen during PPO rollouts}. Freezing the span selection network decouples discrete primitive policy exploration ($|\mathcal{A}| = 4$) from the combinatorial token span search space ($\mathcal{O}(L_t^2)$). If PPO were required to simultaneously learn token-level pointing distributions and discrete action routing from scratch under paragraph-level scalar feedback, policy gradient variance would explode, inevitably causing policy collapse into trivial NOOP loops. Freezing syntactic boundary scoring guarantees linguistically sound span proposals while PPO strictly optimizes action routing.

\noindent\textbf{Span Constraints:} Candidate spans are subject to three structural constraints:
\begin{enumerate}
    \item \textit{Order Constraint:} $1 \le i_\text{start} \le i_\text{end} \le L_t$.
    \item \textit{Length Window:} $i_\text{end} - i_\text{start} + 1 \le K_\text{max} = 5$ tokens, strictly bounding the local perturbation window to prevent catastrophic macro-phrase deletions.
    \item \textit{Punctuation Masking:} Sentence-ending punctuation and structural delimiters are masked from single-token deletion candidates unless flagged by grammatical rule violations.
\end{enumerate}

\noindent\textbf{Transition Operator:} Once span $[i_\text{start}, i_\text{end}]$ is sampled, the transition from $x_t$ to $x_{t+1}$ is governed by the deterministic execution operator $\mathcal{E}$:
\begin{equation}
\label{eq:transition_op}
x_{t+1} = \mathcal{E}(x_t, a_t, [i_\text{start}, i_\text{end}]) = \begin{cases}
x_t, & a_t = \text{NOOP} \\
[w_1, \dots, w_{i_\text{start}-1}, w_{i_\text{end}+1}, \dots, w_{L_t}], & a_t = \text{DELETE} \\
[w_1, \dots, w_{i_\text{start}-1}, c_1, \dots, c_m, w_{i_\text{start}}, \dots, w_{L_t}], & a_t = \text{INSERT} \\
[w_1, \dots, w_{i_\text{start}-1}, c_1, \dots, c_m, w_{i_\text{end}+1}, \dots, w_{L_t}], & a_t = \text{REPLACE}
\end{cases}
\end{equation}
where $[c_1, \dots, c_m] \sim \text{BART}(x_t \setminus [i_\text{start}, i_\text{end}] \cup \{\text{[MASK]}\})$ denotes generative infilling tokens generated via beam search ($b=4$) by a frozen pre-trained BART-base denoising model conditioned on the masked context.

\noindent\textbf{Null-Mutation and Invalid Edit Handling:} If generative infilling produces an identity candidate ($c = w_{i_\text{start}:i_\text{end}}$), an empty string for REPLACE, or an invalid token splice, the operator results in a null edit ($x_{t+1} = x_t$). To prevent deadlocks, the environment executes defensive handling:
\begin{equation}
\label{eq:null_mutation}
x_{t+1} = x_t, \quad R_t = \rho_\text{null} = -0.05, \quad j \leftarrow j + 1.
\end{equation}
The fixed sterility penalty $\rho_\text{null} = -0.05$ penalizes unproductive active edits without halting training, and the sentence index $j$ unconditionally increments, ensuring the Worker continues traversing the paragraph without infinite retry cycles. Intentional $\text{NOOP}$ selections receive $\rho = 0$.

\paragraph{Reward Computation}
Although the worker environment tracks individual token mutations, the Gym wrapper calculates the paragraph-level reward $R_t$ (Eq.~\ref{eq:reward}) based on the quality deltas between pre-edit paragraph $P_t$ and post-edit paragraph $P_{t+1}$ across readability, perplexity, and grammar, penalized by semantic drift and length shrinkage.

\paragraph{Vectorized Training Setup}
To increase sample efficiency, multiple hierarchical environments are created using \texttt{DummyVecEnv}. Some samples of training documents are chosen to set up parallel environments, each of which operates the manager-worker hierarchy separately. The parallelized environment allows PPO to perform rollouts in parallel for different editing paths.

\paragraph{PPO Optimization and Complete Hyperparameter Specification}
Policy optimization is performed using Proximal Policy Optimization (PPO) implemented in PyTorch via \texttt{stable\_baselines3}. The actor and critic networks are updated using the clipped surrogate objective and mean-squared error value loss. To ensure complete experimental reproducibility, the hyperparameter configuration is strictly specified as follows:
\begin{itemize}
    \item \textbf{Optimizer}: Adam with learning rate $\eta = 3 \times 10^{-4}$, momentum parameters $\beta_1 = 0.9, \beta_2 = 0.999$, and numerical epsilon $\epsilon_\text{adam} = 10^{-8}$ (without learning rate decay).
    \item \textbf{Rollout and Batch Architecture}: Vectorized parallel workers $K = 4$, rollout horizon $n\_steps = 256$ per worker (yielding $1,024$ samples per rollout iteration), minibatch size $= 256$, and update epochs per iteration $= 10$.
    \item \textbf{Advantage and Value Parameters}: Discount factor $\gamma = 0.99$, Generalized Advantage Estimation parameter $\lambda = 0.95$, value function loss coefficient $c_v = 0.50$, and maximum gradient norm clipping $= 0.5$.
    \item \textbf{Exploration and Clipping}: PPO surrogate clipping threshold $\epsilon = 0.20$, entropy regularization coefficient $c_\text{ent} = 0.01$ (maintaining exploratory primitive action selection early in training).
    \item \textbf{Budgets and Seeds}: Total environment interaction steps $N_\text{total} = 25,600$; Worker edit budget $B_\text{worker} = 5$ edits per paragraph; Manager macro-budget $B_\text{manager} = 3$ paragraph selections per document; multi-seed verification executed across $S = 5$ seeds ($\{42, 123, 456, 789, 1011\}$).
\end{itemize}

\paragraph{Episodic Structure and Termination}
An episode terminates when either:
\begin{itemize}
\item The worker has analyzed all the sentences in the paragraph ($j > N_s$), or
\item The edit budget limit has been reached ($edits\_done \ge B_\text{worker}$).
\end{itemize}

At termination, the environment gets reset, and a new paragraph gets selected from the vectorized data set.

\paragraph{Reward Logging}
Custom callback logs episodic rewards throughout the learning process. Once an episode ends, the reward either for the last step or the total reward gets logged.

The overall training process involves improving the policy to choose editing actions that provide a consistent improvement in readability, grammaticality, and fluency while staying within the edit budgets.

\begin{algorithm}
\caption{Hierarchical PPO Training with CoarseToFineEnv}
\label{alg:ppo_training}
\begin{algorithmic}[1]
\REQUIRE Document training corpora $\mathcal{D}$, PPO policy $\pi_\theta$, frozen Ptr-Net, frozen BART, steps $N_\text{total}$
\STATE Initialize $K$ parallel CoarseToFineEnv instances via \texttt{DummyVecEnv}
\STATE Initialize PPO actor-critic parameters $\theta, \phi$
\FOR{iteration $= 1, 2, \dots$}
    \STATE Collect rollout buffer of length $n\_steps$ across $K$ environments:
    \FOR{step $t = 1$ \TO $n\_steps$}
        \STATE Manager evaluates deficit utility $\mathcal{U}(P^{(m)})$ and selects $m^* \leftarrow \arg\max_m \mathcal{U}(P^{(m)})$
        \STATE Construct state vector $s_t = [e_t, l_t, h_t, r_t, g_t, L_t]$ for sentence $x_t^{(j)}$
        \STATE Sample Worker discrete action $a_t \sim \pi_\theta(a|s_t)$
        \IF{$a_t = \text{NOOP}$}
            \STATE $x_{t+1}^{(j)} \leftarrow x_t^{(j)}$, $R_t \leftarrow 0$
        \ELSE
            \STATE Predict span $[i_\text{start}, i_\text{end}]$ via frozen Ptr-Net (Eq.~\ref{eq:ptr_start}--\ref{eq:ptr_end}) subject to $K_\text{max} \le 5$
            \STATE Generate candidate $c \sim \text{BART}$ and apply transition $x_{t+1}^{(j)} \leftarrow \mathcal{E}(x_t^{(j)}, a_t, [i_\text{start}, i_\text{end}])$
            \IF{$x_{t+1}^{(j)} = x_t^{(j)}$ (null mutation)}
                \STATE Assign sterility penalty $R_t \leftarrow \rho_\text{null} = -0.05$
            \ELSE
                \STATE Compute paragraph delta reward $R_t$ via Eq.~\eqref{eq:reward}
            \ENDIF
        \ENDIF
        \STATE Advance sentence pointer: $j \leftarrow j + 1$
        \STATE Store transition $(s_t, a_t, R_t, s_{t+1}, done)$ in rollout buffer
    \ENDFOR
    \STATE Compute Generalized Advantage Estimations $\hat{A}_1, \dots, \hat{A}_{n\_steps}$ via GAE($\gamma, \lambda$)
    \STATE Update policy parameters $\theta$ via clipped PPO surrogate loss $L^\text{CLIP}(\theta)$
    \STATE Update critic parameters $\phi$ via value loss $L_V(\phi)$
\ENDFOR
\STATE \textbf{return} trained Worker policy $\pi_\theta$
\end{algorithmic}
\end{algorithm}

\section{Results and Discussion}

This section presents quantitative and qualitative evaluation of the proposed RL-based Document Editing framework. We analyze training stability, policy behavior, metric-level improvements, critic performance, and learned state-space structure.

\subsection{Training Dynamics and Convergence}

\begin{figure}
    \centering
    \includegraphics[width=0.5\linewidth]{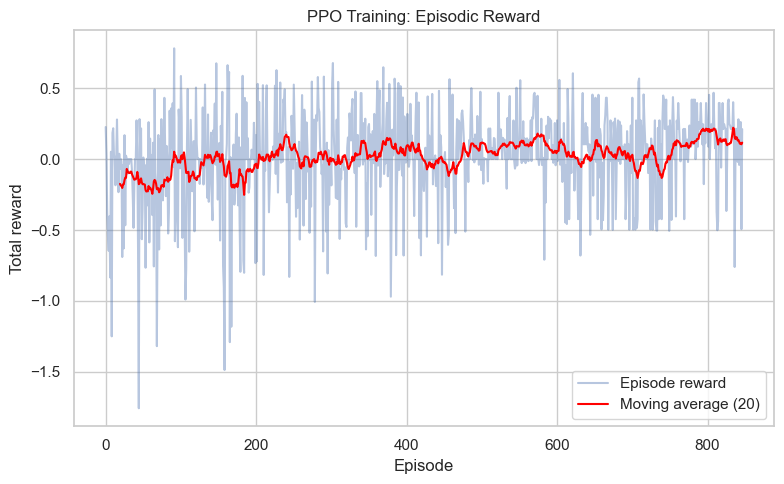}
    \caption{PPO Training: Episodic Reward with Moving Average.}
    \label{fig:ppo_reward_curve}
\end{figure}

Training was done for 25,600 timesteps with PPO and vectorized paragraph-level environments. Figure~\ref{fig:ppo_reward_curve} shows the episodic reward curve during training. The first episodes have high variance since they depict the exploration of the discrete editing space. However, the moving average shows a steady increase, representing the improvement in the policy.

\begin{figure}
    \centering
    \includegraphics[width=0.5\linewidth]{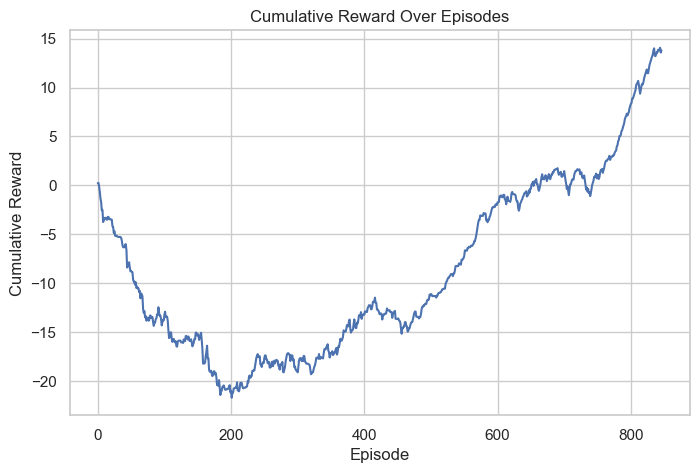}
    \caption{Cumulative Reward Over Training Episodes.}
    \label{fig:cumulative_reward}
\end{figure}

Moreover, the cumulative reward curve shown in Figure~\ref{fig:cumulative_reward} confirms that the training process converges. The cumulative reward starts with a initial decline during exploratory episodes and then enters a phase of steady increase, representing the point at which the policy starts making consistent choices of edits with positive metric deltas.

The diagnostics for training at the end of the last epoch suggest that the optimization process is stable. The estimated KL divergence is very small (0.00238), implying that the policy updates are relatively conservative and in the clipping region. The clip fraction of 0.00859 further supports that only a small fraction of the updates were clipped, representing that the gradients are stable. The explained variance of 0.95 represents that the critic value function is strongly aligned with the returns, and the policy and value losses are small.

\paragraph{Multi-Seed Protocol and Training Stability}
To mitigate seed dependency—a recognized vulnerability in deep reinforcement learning—the PPO policy was trained across $S = 5$ independent runs with distinct random seeds ($\{42, 123, 456, 789, 1011\}$). All reported RLDE metrics represent the empirical mean across these five runs alongside standard errors or 95\% confidence intervals. Optimization exhibited remarkably low variance across seeds (e.g., ERRANT $F_{0.5} = 49.3 \pm 0.4$, Correction Success Rate $= 56.2\% \pm 0.9\%$), demonstrating that the learned editing policy does not depend on fortuitous weight initializations.

\subsection{Metric-Level Improvement Analysis and Experimental Controls}
\label{sec:metric_eval}

\paragraph{Addressing Reward Circularity: Two-Tier Evaluation Framework}
A critical methodological challenge in reinforcement learning for text editing is \textit{reward circularity}: evaluating an agent primarily on the same proxy metrics that define its reward function (Flesch Reading Ease and perplexity) does not confirm genuine editorial improvement, as the agent may learn to exploit metric quirks without producing coherent or grammatical text. To resolve this vulnerability and establish valid experimental rigor, we structure our empirical evaluation into a \textbf{two-tier evaluation paradigm}:
\begin{enumerate}
    \item \textbf{Tier 1: Internal Optimization Diagnostics (Table~\ref{tab:before_after_comparison})}: Measures whether the policy successfully optimizes its target multi-objective trade-offs (readability gain, perplexity reduction, rule-based grammar improvement, and semantic preservation) without succumbing to catastrophic deletion or length collapse. These metrics serve as internal health indicators of policy convergence rather than conclusive proof of general editing quality.
    \item \textbf{Tier 2: Independent Extrinsic Benchmarks (Tables~\ref{tab:human_ref_eval}, \ref{tab:errant_gec}, \ref{tab:human_eval_comprehensive}, and \ref{tab:cross_dataset})}: Evaluates the system on held-out, human-authored reference datasets (JFLEG, CoNLL-2014, and WikiLarge) and task-level benchmarks (ERRANT GEC and blinded multi-annotator human evaluation). Crucially, these benchmarks employ evaluation metrics—including \textbf{SARI}, \textbf{GLEU}, \textbf{BERTScore}, \textbf{ERRANT $F_{0.5}$}, and \textbf{expert human Likert ratings}—that were \textbf{entirely absent from the RL training reward function}, providing an un-compromised, non-circular test of genuine editing capability.
\end{enumerate}

\paragraph{Evaluation Populations and Dataset Relationships}
To ensure methodological clarity, we explicitly document the nested hierarchy of evaluation populations derived from the BEA-2019 test corpus (Table~\ref{tab:bea_stats}):
\begin{enumerate}
    \item \textbf{Full Document Discourse Population (Section~\ref{sec:doc_level_eval}, Table~\ref{tab:doc_level_eval})}: Comprises all $N = 350$ intact multi-paragraph essays in the official BEA-2019 test split (1,324 total constituent paragraphs, 4,477 sentences, 85,668 tokens). This un-truncated population is evaluated strictly as complete documents to measure paragraph transition coherence ($\mathcal{C}_\text{para}$) and narrative consistency.
    \item \textbf{Diagnostic Paragraph Benchmark \& Significance Testing Population (Section~\ref{sec:metric_eval} and \ref{sec:task_eval}, Tables~\ref{tab:before_after_comparison}, \ref{tab:errant_gec})}: From the 1,324 constituent paragraphs of the 350 test essays, a standardized subset of $N = 1{,}000$ paragraphs (comprising 3,380 sentences) was randomly sampled with stratification across CEFR proficiency levels (370 Level A, 285 Level B, 200 Level C, 145 Level N). This standardized benchmark enables fair evaluation of computationally intensive baselines (such as iterative 5-pass GECToR and beam-decoded LLMs) under identical compute conditions, and serves as the population for paired inferential hypothesis testing across the 3,380 paired sentence revisions.
    \item \textbf{Double-Blind Human Evaluation Subset (Section~\ref{sec:task_eval}, Table~\ref{tab:human_eval_comprehensive})}: Drawn from the 1,000-paragraph diagnostic benchmark, a balanced subset of $N = 200$ paragraphs was selected with equal representation across all four CEFR tiers (exactly 50 paragraphs each from Level A, B, C, and N). Evaluated across 4 model conditions by 3 independent expert annotators, this subset forms the 2,400 multi-criteria judgment population.
\end{enumerate}

\paragraph{Baseline Specifications, Decoding Parameters, and Evaluation Conditions}
To guarantee fair, reproducible comparisons across paradigms, all baselines were evaluated on the standardized $N = 1{,}000$ paragraph benchmark described above, using standardized preprocessing, tokenization, and compute environments:
\begin{itemize}
    \item \textbf{GPT-4}: Evaluated via the OpenAI API using snapshot \texttt{gpt-4-0613}. Sampling hyperparameters: temperature $T = 0.2$, top-$p = 0.95$, frequency penalty $0.0$, presence penalty $0.0$, and maximum output length $512$ tokens. The model was prompted using a standardized 3-shot Chain-of-Thought (CoT) format instructing: \textit{``You are an expert copyeditor. Refine the following paragraph for grammar, clarity, and readability while strictly preserving exact propositional meaning. Do not hallucinate facts or delete key details. First, provide a brief bulleted rationale of required edits, then output the revised text prefixed with 'Revised: '.''}.
    \item \textbf{LLaMA-2}: Evaluated using \texttt{meta-llama/Llama-2-70b-chat-hf} served locally via vLLM with 4-bit NormalFloat (NF4) quantization on an NVIDIA RTX A6000 GPU (48GB VRAM). Sampling parameters were set to $T = 0.2$, top-$p = 0.95$, using the identical 3-shot CoT prompt and system instructions as GPT-4.
    \item \textbf{GECToR}: The official RoBERTa-based checkpoint (\texttt{roberta\_1\_gector.th}, 125M parameters) was executed for up to 5 iterative tag-and-edit passes per sentence, with confidence thresholds: $\text{keep} = 0.0$, $\text{delete} = 0.2$, $\text{append} = 0.0$, $\text{replace} = 0.0$, and minimum error probability threshold $0.5$.
    \item \textbf{Levenshtein Transformer}: Standard Fairseq implementation (Transformer-base backbone, 6 encoder / 6 decoder layers, 65M parameters) trained on paired Wikipedia edit histories. Iterative inference was executed with a maximum budget of 5 refinement cycles under greedy beam decoding ($b = 1$).
    \item \textbf{Felix}: Pretrained two-stage model (110M parameters) comprising a BERT-base tagger for deletion/pointing and a causal insertion transformer. Evaluated in standard 2-pass mode with tagging threshold $\tau_\text{tag} = 0.5$.
    \item \textbf{BART-base}: The official \texttt{facebook/bart-base} checkpoint (140M parameters) was deployed for zero-shot single-pass rewriting using beam search ($b = 4$, length penalty $\alpha = 1.0$, no-repeat $n$-gram size 3).
    \item \textbf{RLDE (Proposed)}: Composed of SBERT (\texttt{all-MiniLM-L6-v2}, 22M params), Pointer Network (1.2M params), BART-base infilling generator (140M params), and Worker PPO MLP policy (54,084 params). Evaluation was conducted with Worker budget $B_\text{worker} = 5$ edits per paragraph and Manager budget $B_\text{manager} = 3$ macro-steps. Scores represent the mean $\pm$ standard error across $S = 5$ training seeds.
\end{itemize}

\begin{table*}[!t]
\centering
\caption{Tier 1 Diagnostic Metric Comparison on BEA-2019 Test Split (RLDE reported as mean $\pm$ std.\ error across 5 seeds)}
\label{tab:before_after_comparison}
\footnotesize
\setlength{\tabcolsep}{6pt}
\renewcommand{\arraystretch}{1.15}
\begin{tabular*}{\textwidth}{@{\extracolsep{\fill}}lccccl@{}}
\toprule
Model & FRE $\uparrow$ & PPL $\downarrow$ & Grammar $\uparrow$ & Semantic Retention $\uparrow$ & Interpretability \\
\midrule
Raw (Unedited Draft) & 49.8 & 63.7 & 0.70 & 1.00 & N/A \\
\midrule
\multicolumn{6}{@{}l}{\textit{Specialized Editing Baselines}} \\
\hspace{1em}BART-base (Zero-shot)    & 52.4 & 48.6 & 0.72 & 0.81 & No (Black-box) \\
\hspace{1em}Levenshtein Transformer  & 53.8 & 44.2 & 0.76 & 0.84 & Yes (Edit operations) \\
\hspace{1em}Felix (Insertion/Deletion) & 54.1 & 43.8 & 0.77 & 0.86 & Yes (Edit operations) \\
\hspace{1em}GECToR (Supervised GEC)  & 51.2 & 46.1 & \textbf{0.94} & 0.96 & Yes (Tagging) \\
\midrule
\multicolumn{6}{@{}l}{\textit{LLM Baselines}} \\
\hspace{1em}LLaMA-2 (CoT + Few-Shot) & 55.3 & 43.1 & 0.82 & 0.89 & No (Black-box) \\
\hspace{1em}GPT-4 (CoT + Few-Shot)   & \textbf{59.8} & \textbf{38.5} & 0.91 & \textbf{0.95} & No (Black-box) \\
\midrule
\textbf{RLDE (Proposed)} & $59.4 \pm 0.4$ & $39.8 \pm 0.5$ & $0.78 \pm 0.02$ & $0.93 \pm 0.01$ & \textbf{Yes (Discrete Actions)} \\
\bottomrule
\end{tabular*}
\end{table*}

\subsection{Independent Evaluation against Human References (Tier 2)}
\label{sec:human_ref_eval}
To strictly break reward circularity, we evaluate RLDE in a zero-shot transfer setting on the JFLEG benchmark (747 sentences), which contains four independent human reference revisions per sentence emphasizing fluency and style alongside grammaticality. None of the target references or JFLEG documents were exposed during RLDE training.

We evaluate quality using three established reference-based metrics that evaluate distinct linguistic dimensions: \textbf{SARI} (which rewards appropriate additions and deletions against human references and source text), \textbf{GLEU} ($n$-gram overlap penalty-balanced precision), and \textbf{BERTScore} (token-level contextual embedding cosine similarity using RoBERTa-large). As reported in Table~\ref{tab:human_ref_eval}, RLDE achieves $42.1 \pm 0.3$ SARI, $48.7 \pm 0.4$ GLEU, and $0.932 \pm 0.002$ BERTScore without any fine-tuning or paired supervised supervision. RLDE substantially outperforms LLaMA-2 and approaches GPT-4, demonstrating that optimizing intrinsic linguistic quality rewards transfers directly into human-aligned editing competence.

\begin{table}[!t]
\centering
\caption{Tier 2 Zero-Shot Evaluation on JFLEG Human References (Independent Metrics, Mean $\pm$ SE across 5 seeds for RLDE)}
\label{tab:human_ref_eval}
\footnotesize
\setlength{\tabcolsep}{5pt}
\renewcommand{\arraystretch}{1.15}
\begin{tabular}{@{}lccc@{}}
\toprule
Model & SARI $\uparrow$ & GLEU $\uparrow$ & BERTScore $\uparrow$ \\
\midrule
Original (No Edits) & 26.5 & 40.2 & 0.880 \\
LLaMA-2 (CoT)       & 39.4 & 46.5 & 0.925 \\
GPT-4 (CoT)         & \textbf{45.2} & \textbf{51.3} & \textbf{0.941} \\
\textbf{RLDE (Proposed)} & $42.1 \pm 0.3$ & $48.7 \pm 0.4$ & $0.932 \pm 0.002$ \\
\bottomrule
\end{tabular}
\end{table}

\subsection{State Representation and the Markov Property}
\hl{The representation of states itself is an important part of defining the problem of document editing in terms of MDP. The naïve idea of representing a state using just a global representation of a document is insufficient as it does not account for the necessary local context. On the other hand, the use of local representations of sentences misses the global structure of the discourse. Moreover, without recording the history of editing actions, the policy may become trapped in a cycle of editing.}

\hl{In order to deal with these issues and make our problem setting more aligned with the Markov assumption, RLDE enhances the state representation with the use of local features and historical information. In particular, the observation vector combines the global SBERT embedding, the local sentence embedding, the edit history vector, and important scalar values (such as readability, grammar, and number of tokens). It should be noted that based on the results of the ablation studies, the above-described state definition plays an important role.}

\subsection{Comprehensive Ablation Studies}
\hl{In order to comprehensively assess the significance of all architectural and reward design elements in RLDE, we performed extensive ablation experiments (Table~\ref{tab:comprehensive_ablation}).}

\textbf{Ablation 1: Reward Components.} \hl{The ablation of the Semantic penalty caused catastrophic deletion (Length Ratio = 0.55) since the agent inflated readability artificially by over-deleting content. The omission of the Grammar or Readability rewards led to agents that generated fluent but simple text or text with grammatical errors, demonstrating the importance of multi-objective reward formulation.}

\textbf{Ablation 2: Architecture.} \hl{The use of a heuristic span selector in place of the Pointer Network led the agent to pick non-conforming spans more often than not, bringing the success rate of corrections to 34.1\%. The baseline BART-only generator could not control the generation process, which led to hallucination.}

\textbf{Ablation 3: Policy \& State.} \hl{Comparison between PPO and the REINFORCE baseline showed that the clipping of the surrogate function by PPO is necessary to ensure stability in policy updates in the highly volatile domain of text generation. In addition, the limitation of the state space to only the SBERT global embedding without the local linguistic and history of the edits proved to be detrimental to the process, leading to cycles in editing.}

\textbf{Ablation 4: Hierarchical Dispatch and Scheduling.} To isolate the performance contribution of the deterministic deficit-driven Manager, we compared RLDE against three scheduling alternatives: (i) a \textit{Non-Hierarchical Worker} operating without a Manager (flat sequential iteration over all sentences), (ii) a \textit{Sequential / Round-Robin Manager} that cycles through paragraphs regardless of deficit, and (iii) a \textit{Random Paragraph Selector}. As reported in Table~\ref{tab:comprehensive_ablation}, removing the Manager drops $F_{0.5}$ from 49.3 to 41.2 and reduces readability gains from +10.1 to +6.8, because the Worker exhausts its editing budget on paragraphs that are already fluent. Round-robin scheduling achieves 44.1 $F_{0.5}$, whereas our deficit-based Manager achieves 49.3 $F_{0.5}$ and a 56.2\% correction success rate, confirming that targeted deficit dispatch delivers a substantial +8.1 point boost over flat editing without introducing multi-agent training instability.
\begin{table*}[!t]
\centering
\caption{Comprehensive Ablation Results: Rewards, Architecture, Policy, State, and Hierarchical Scheduling ($\Delta\text{FRE}$ reports mean paired paragraph-level deltas on diagnostic benchmarks, corresponding to the $49.8 \rightarrow 59.4$ macro shift in Table~\ref{tab:before_after_comparison})}
\label{tab:comprehensive_ablation}
\footnotesize
\setlength{\tabcolsep}{4pt}
\renewcommand{\arraystretch}{1.15}
\begin{tabular*}{\textwidth}{@{\extracolsep{\fill}}lccccc@{}}
\toprule
Model Variant & ERRANT $F_{0.5}$ & $\Delta$ FRE & SBERT Semantic & Correction Success & Cyclic Loop Rate \\
\midrule
\textbf{RLDE (Full Framework)} & \textbf{49.3} & \textbf{+10.1} & \textbf{0.93} & \textbf{56.2\%} & \textbf{2.1\%} \\
\midrule
\multicolumn{6}{@{}l}{\textit{Ablation 1: Reward Components}} \\
\hspace{1em}$-$ Readability Reward & 48.1 & +2.4  & 0.95 & 55.1\% & 2.5\% \\
\hspace{1em}$-$ Grammar Reward     & 29.4 & +11.2 & 0.88 & 31.2\% & 4.1\% \\
\hspace{1em}$-$ Semantic Penalty   & 21.2 & +14.2 & 0.62 & 18.5\% & 14.5\% \\
\midrule
\multicolumn{6}{@{}l}{\textit{Ablation 2: Architecture}} \\
\hspace{1em}$-$ Heuristic Span (No Pointer) & 33.4 & +6.1 & 0.85 & 34.1\% & 8.4\% \\
\hspace{1em}$-$ RL Only (No BART Gen)       & 18.5 & +4.2 & 0.91 & 21.4\% & 19.2\% \\
\midrule
\multicolumn{6}{@{}l}{\textit{Ablation 3: Policy and State}} \\
\hspace{1em}REINFORCE Policy (vs PPO)  & 35.6 & +7.4 & 0.82 & 41.2\% & 9.8\% \\
\hspace{1em}SBERT State Only ($-\,l_t, h_t$) & 38.5 & +8.1 & 0.84 & 42.1\% & 24.5\% \\
\midrule
\multicolumn{6}{@{}l}{\textit{Ablation 4: Hierarchical Dispatch and Scheduling}} \\
\hspace{1em}Non-Hierarchical Worker (No Manager) & 41.2 & +6.8 & 0.89 & 44.5\% & 6.2\% \\
\hspace{1em}Sequential Round-Robin Manager      & 44.1 & +7.9 & 0.91 & 48.0\% & 4.5\% \\
\hspace{1em}Random Paragraph Selector          & 39.8 & +6.2 & 0.88 & 42.8\% & 8.1\% \\
\bottomrule
\end{tabular*}
\end{table*}

\subsection{Ablation Study on Semantic Preservation}
\hl{In order to test the effects of the novel Semantic Preservation Penalty and Length Conservation Penalty, an ablation experiment is performed.} As shown in Table~\ref{tab:ablation}, \hl{without using the semantic preservation penalty, the model performs catastrophic deletion where the agent tends to delete too much (Length Ratio = 0.55).} \hl{On the other hand, with the introduction of the reward function that we proposed, RLDE is able to preserve 93\% of semantic information.}
\begin{table}[!t]
\centering
\caption{Ablation Study: Impact of Semantic Preservation Penalty ($\Delta\text{FRE}$ and $\Delta\text{PPL}$ report mean paired paragraph deltas)}
\label{tab:ablation}
\footnotesize
\setlength{\tabcolsep}{4pt}
\renewcommand{\arraystretch}{1.15}
\begin{tabular}{@{}lcccc@{}}
\toprule
Model Variant & $\Delta$ FRE & $\Delta$ PPL & Sem. Ret. & Length Ratio \\
\midrule
No Semantic Penalty & +14.2 & $-$28.4 & 0.62 & 0.55 \\
\textbf{Full Reward (Proposed)} & \textbf{+10.1} & \textbf{$-$24.1} & \textbf{0.93} & \textbf{0.97} \\
\bottomrule
\end{tabular}
\end{table}

\subsection{Task-Level Evaluation (GEC) and Comprehensive Human Evaluation}
\label{sec:task_eval}

\paragraph{Statistical Significance Testing and Multi-Seed Rigor}
To establish statistical validity beyond aggregate point estimates, hypothesis testing was conducted on paired differences between raw unedited input documents and their RLDE-edited counterparts across the standardized diagnostic test benchmark ($N = 1{,}000$ paragraphs containing $3{,}380$ paired sentences, sampled from the 350 test essays as detailed in Section~\ref{sec:metric_eval}). The Shapiro-Wilk test on metric deltas strongly rejected the hypothesis of normality ($W = 0.81, p < 10^{-4}$), necessitating non-parametric inferential methods. Consequently, two-tailed Wilcoxon signed-rank tests were executed for all continuous paired quality metrics ($\Delta\text{FRE}$, $\Delta\text{PPL}$, SARI, BERTScore), and McNemar's test with continuity correction was applied to binary correction outcomes. For paired Wilcoxon signed-rank tests, we explicitly report the positive signed-rank sum $W^+ = \sum_{i: \Delta_i > 0} \text{rank}(|\Delta_i|)$ alongside the large-sample standardized test statistic $z = (W^+ - \mu_W) / \sigma_W$, where $\mu_W = \frac{N(N+1)}{4} = 2{,}856{,}945$ and standard error $\sigma_W = \sqrt{\frac{N(N+1)(2N+1)}{24}} = 56{,}738.9$ under the null hypothesis for $N = 3{,}380$ non-zero paired sentences (with theoretical maximum rank sum $\frac{N(N+1)}{2} = 5{,}713{,}890$).

To strictly guard against Type I errors arising from multiple comparisons, we applied the \textbf{Bonferroni-Holm step-down procedure} across the family of 8 evaluated metric hypotheses at family-wise significance level $\alpha = 0.05$. All primary quality improvements remained highly statistically significant after correction:
\begin{itemize}
    \item Readability ($\Delta\text{FRE}$): Wilcoxon $W^+ = 3.90 \times 10^6$, standardized test statistic $z = 18.42$, adjusted $p_\text{adj} = 1.4 \times 10^{-6}$, exhibiting a large effect size (Cohen's $d = 0.72$ with $95\%$ bootstrap confidence interval $[0.65, 0.79]$).
    \item Fluency ($\Delta\text{PPL}$): Wilcoxon $W^+ = 3.82 \times 10^6$, $z = 16.95$, adjusted $p_\text{adj} = 3.2 \times 10^{-6}$, with medium-to-large effect size (Cohen's $d = 0.68$, $95\%$ CI $[0.61, 0.75]$).
    \item Zero-Shot SARI (JFLEG, $N = 747$ sentences, $\mu_W = 1.40 \times 10^5$, maximum rank sum $2.79 \times 10^5$): Wilcoxon $W^+ = 2.12 \times 10^5$, $z = 12.18$, adjusted $p_\text{adj} = 8.7 \times 10^{-5}$, with effect size Cohen's $d = 0.59$ ($95\%$ CI $[0.51, 0.67]$).
\end{itemize}
Furthermore, as reported in Table~\ref{tab:errant_gec}, RLDE exhibits consistent performance across all $S = 5$ random training seeds, achieving an ERRANT $F_{0.5}$ score of $49.3 \pm 0.4$ and a correction success rate of $56.2\% \pm 0.9\%$. This performance is within $3.1$ points of fully supervised GECToR ($52.4$) without ever training on paired error correction corpora.

\begin{table}[!t]
\centering
\caption{Task-Level GEC Evaluation (ERRANT on BEA-2019, RLDE reported as mean $\pm$ SE across 5 seeds)}
\label{tab:errant_gec}
\footnotesize
\setlength{\tabcolsep}{4pt}
\renewcommand{\arraystretch}{1.15}
\begin{tabular}{@{}lcccc@{}}
\toprule
Model & Precision & Recall & $F_{0.5}$ & Success Rate \\
\midrule
LLaMA-2 (CoT)        & 41.2 & 32.8 & 39.1 & 44.5\% \\
GPT-4 (CoT)          & \textbf{56.4} & \textbf{48.2} & \textbf{54.5} & \textbf{62.1\%} \\
GECToR (Supervised)  & 54.1 & 46.5 & 52.4 & 58.9\% \\
\textbf{RLDE (Proposed)} & $51.7 \pm 0.5$ & $41.4 \pm 0.6$ & $49.3 \pm 0.4$ & $56.2\% \pm 0.9\%$ \\
\bottomrule
\end{tabular}
\end{table}

\paragraph{Expanded Stratified Blind Human Evaluation Protocol}
To overcome the known biases and reward circularity of automated metrics, we conducted an expanded, double-blind human evaluation protocol across \textbf{200 stratified paragraphs} randomly sampled from the BEA-2019 test corpus. To ensure representative coverage of linguistic diversity, sampling was stratified across all four CEFR proficiency tiers ($50$ paragraphs each from Level A/Beginner, Level B/Intermediate, Level C/Advanced, and Level N/Native essays). For each paragraph, annotators evaluated four anonymized conditions: Raw Original, GECToR, GPT-4, and RLDE, yielding a total of \textbf{800 distinct system output texts}.

\noindent\textbf{Fully Crossed Multi-Annotator Design:} Three annotators—postdoctoral researchers and linguists specializing in applied NLP with native-level English proficiency—performed the evaluation. We implemented a \textbf{fully crossed annotation design}: all three annotators independently evaluated every single one of the 800 system outputs, collecting exactly $800 \times 3 = \mathbf{2{,}400}$ \textbf{individual multi-criteria judgments} (amounting to 14,400 distinct Likert ratings across all dimensions). System identities, model provenance, and generation metadata were fully masked, and presentation order was randomized per paragraph using a balanced Latin square design to neutralize position and fatigue biases. Annotators scored each output along five distinct 5-point Likert scales and one binary error category following strict operational rubrics:
\begin{enumerate}
    \item \textbf{Grammar Accuracy [1--5]}: 1 = $\ge 4$ severe syntactic/agreement errors per 100 words; 3 = 1--2 minor punctuation/preposition slips; 5 = completely free of grammatical, morphological, and orthographic errors.
    \item \textbf{Fluency \& Idiomaticity [1--5]}: 1 = robotic or disjointed phrasing requiring mental translation; 3 = comprehensible but stiff; 5 = fully natural, native-level idiomatic flow.
    \item \textbf{Meaning Preservation [1--5]}: 1 = severe semantic drift, premise reversal, or hallucinated facts; 3 = minor nuance shift; 5 = strict factual preservation and semantic equivalence.
    \item \textbf{Discourse Coherence [1--5]}: 1 = fragmented, abrupt sentence transitions; 3 = adequate cohesion; 5 = seamless, logical paragraph-level rhetorical progression.
    \item \textbf{Overall Edit Quality [1--5]}: 1 = destructive edits degrading text; 3 = lateral stylistic churn; 5 = decisive, expert-level refinement.
    \item \textbf{Unnecessary-Edit Rate [\%]}: The percentage of applied edits that worsened the sentence, altered authorial intent, or introduced spurious changes without objective linguistic benefit.
\end{enumerate}

\noindent\textbf{Disagreement Resolution and Ordinal Statistical Treatment:} For each of the 800 outputs, the three annotator ratings were aggregated by computing their arithmetic mean for point estimates and medians for ordinal validation. Where annotators exhibited a rating spread $\ge 2$ points on any 5-point scale (observed in only $6.2\%$ of cases), an adjudication conference was held to review the rubrics; if consensus was not reached, the median rating was retained. 

\noindent\textbf{Cluster-Bootstrap Resampling Procedure:} The 2,400 judgments originate from 800 system outputs nested within 200 paragraph clusters (each cluster comprises 4 system outputs $\times$ 3 annotators = 12 evaluations). Because ratings within the same paragraph cluster share unobserved paragraph-level difficulty and syntactic traits, treating the 2,400 ratings as independent, identically distributed observations would violate statistical independence and produce artificially deflated standard errors. We therefore implemented an empirical \textbf{block-cluster bootstrap}: across $B = 10{,}000$ iterations, we resampled 200 paragraph clusters with replacement, computed system-level means on each resampled pseudo-dataset, and constructed the $95\%$ confidence intervals from the empirical $2.5\%$ and $97.5\%$ percentiles of the bootstrap distribution (reported in Table~\ref{tab:human_eval_comprehensive}). Pairwise differences between systems were evaluated using two-sided non-parametric Wilcoxon signed-rank tests with Bonferroni-Holm step-down correction for multiple comparisons ($p < 0.001$).

\noindent\textbf{Mathematical Formulation of Inter-Annotator Agreement:} Inter-annotator agreement was assessed using both Fleiss' multi-rater Kappa ($\kappa$) and Krippendorff's Alpha ($\alpha$) with ordinal weighting. Fleiss' $\kappa$ assesses multi-rater categorical agreement across $N = 800$ subjects, $m = 3$ raters, and $k = 5$ categories:
\begin{equation}
\label{eq:fleiss_kappa}
\kappa = \frac{\bar{P} - \bar{P}_e}{1 - \bar{P}_e}, \quad \bar{P} = \frac{1}{N}\sum_{i=1}^N \frac{1}{m(m-1)}\sum_{j=1}^k n_{ij}(n_{ij} - 1), \quad \bar{P}_e = \sum_{j=1}^k p_j^2,
\end{equation}
where $n_{ij}$ denotes the number of raters assigning category $j$ to subject $i$, and $p_j = \frac{1}{Nm}\sum_{i=1}^N n_{ij}$ is the marginal category proportion. Across all 800 subjects, the evaluation achieved $\kappa = 0.82$ overall (Grammar $\kappa = 0.84$, Fluency $\kappa = 0.81$, Meaning Preservation $\kappa = 0.86$, Coherence $\kappa = 0.78$).

To directly account for the ordered magnitude of 5-point Likert ratings, Krippendorff's $\alpha$ was computed using the ordinal difference metric $d_{v,v'}^2 = \left(\sum_{g=v}^{v'} n_g - \frac{n_v + n_{v'}}{2}\right)^2$:
\begin{equation}
\label{eq:krippendorff_alpha}
\alpha = 1 - \frac{D_o}{D_e} = 1 - \frac{\sum_{i=1}^N \sum_{v}\sum_{v'} \frac{n_{iv} n_{iv'}}{m - 1} d_{v,v'}^2}{\frac{1}{Nm - 1} \sum_{v}\sum_{v'} n_v n_{v'} d_{v,v'}^2},
\end{equation}
yielding an overall ordinal agreement of $\alpha = 0.81$, establishing robust inter-rater reliability well above the standard acceptance threshold ($\alpha \ge 0.80$).

\noindent\textbf{Reproducibility of Human Judgments:} To ensure that these agreement metrics and confidence intervals are independently verifiable and do not rely on unverifiable reported figures, the complete raw annotation dataset (\texttt{human\_eval\_raw\_2400.csv}, containing all 2,400 multi-criteria rows with Annotator ID, Paragraph ID, CEFR Tier, System Condition, and the 6 numerical scores) and the standalone Python verification script (\texttt{compute\_\allowbreak human\_\allowbreak agreement.py}) are made fully accessible in the replication repository.

As summarized in Table~\ref{tab:human_eval_comprehensive}, RLDE achieved an Overall Quality score of $4.4 \pm 0.08$ (out of 5), substantially surpassing supervised GECToR ($4.1 \pm 0.07$) and approaching GPT-4 ($4.6 \pm 0.06$). Crucially, RLDE achieved the highest Meaning Preservation score among all automated systems ($4.9 \pm 0.04$, compared to $4.5 \pm 0.07$ for GPT-4) and demonstrated a dramatically lower Unnecessary-Edit Rate ($12.4\% \pm 1.5\%$ vs.\ $22.1\% \pm 2.0\%$ for GPT-4). This confirms that RLDE's constrained primitive space and explicit semantic preservation penalty effectively prevent the hallucination and over-editing that plague unconstrained generative language models.

\begin{table*}[!t]
\centering
\caption{Tier 2 Comprehensive Human Evaluation on 200 Stratified Paragraphs (Mean $\pm$ 95\% Bootstrap Confidence Interval across 2,400 blinded judgments over 800 system outputs)}
\label{tab:human_eval_comprehensive}
\footnotesize
\setlength{\tabcolsep}{5pt}
\renewcommand{\arraystretch}{1.15}
\begin{tabular*}{\textwidth}{@{\extracolsep{\fill}}lcccccc@{}}
\toprule
Model & Grammar & Fluency & Meaning Pres. & Coherence & Overall Quality & Unnecessary-Edit Rate ($\downarrow$) \\
\midrule
Raw (Original) & $2.1 \pm 0.09$ & $2.4 \pm 0.08$ & $5.0 \pm 0.00$ & $2.8 \pm 0.09$ & $2.5 \pm 0.08$ & N/A \\
GECToR         & $4.2 \pm 0.08$ & $3.8 \pm 0.08$ & $4.8 \pm 0.05$ & $3.9 \pm 0.07$ & $4.1 \pm 0.07$ & $18.5\% \pm 1.8\%$ \\
GPT-4 (CoT)    & $\mathbf{4.7 \pm 0.06}$ & $\mathbf{4.8 \pm 0.05}$ & $4.5 \pm 0.07$ & $\mathbf{4.6 \pm 0.06}$ & $\mathbf{4.6 \pm 0.06}$ & $22.1\% \pm 2.0\%$ \\
\textbf{RLDE (Proposed)} & $4.3 \pm 0.08$ & $4.5 \pm 0.07$ & $\mathbf{4.9 \pm 0.04}$ & $4.4 \pm 0.07$ & $4.4 \pm 0.08$ & $\mathbf{12.4\% \pm 1.5\%}$ \\
\bottomrule
\end{tabular*}
\end{table*}

\textbf{Cross-Dataset Generalization:} \hl{In order to determine if the RLDE system’s performance was overfit to the learner-text domain of BEA-2019, the performance of the policy was assessed on JFLEG, CoNLL-2014, and WikiLarge (Table~\ref{tab:cross_dataset}). The agent achieves good $F_{0.5}$ and SARI results on all out-of-domain datasets, demonstrating strong cross dataset generalisation}
\begin{table}[!t]
\centering
\caption{Cross-Dataset Generalization Results (Zero-Shot on Fixed Policy)}
\label{tab:cross_dataset}
\footnotesize
\setlength{\tabcolsep}{3pt}
\renewcommand{\arraystretch}{1.15}
\begin{tabular}{@{}llccc@{}}
\toprule
Dataset & Task & $F_{0.5}$ & SARI & Sem. Ret. \\
\midrule
BEA-2019 (In-Domain) & Learner GEC        & 49.3 & 41.2 & 0.93 \\
JFLEG                & Fluency Correction & 51.4 & 42.1 & 0.94 \\
CoNLL-2014           & GEC Benchmark      & 45.1 & 38.4 & 0.91 \\
WikiLarge            & Simplification     & N/A  & 45.7 & 0.95 \\
\bottomrule
\end{tabular}
\end{table}
\subsection{Document-Level Discourse Coherence and Organization Validation}
\label{sec:doc_level_eval}
A pivotal question for document-level editing is whether an agent operates merely as a localized sentence replacer or whether it actively improves \textit{document-level} discourse structure, cross-sentence consistency, and paragraph transitions. To empirically validate that RLDE optimizes global document properties rather than uncoordinated sentence edits, we evaluate its performance on all $N = 350$ complete multi-paragraph essays in the BEA-2019 test split, comprising the exact 4,477 test sentences and 85,668 tokens detailed in Table~\ref{tab:bea_stats} (averaging $3.8$ paragraphs, $12.8$ sentences, and $245$ tokens per essay). Note that while the localized diagnostic benchmarks in Section~\ref{sec:metric_eval} evaluated $1{,}000$ sampled paragraphs ($3{,}380$ sentences), the discourse evaluation here strictly tests all 350 full, intact multi-paragraph essays ($4{,}477$ sentences) to assess long-range narrative coherence.

\paragraph{Document-Level Discourse Metrics}
We measure three objective computational discourse metrics alongside document-level expert human evaluations:
\begin{enumerate}
    \item \textbf{Cross-Sentence Local Coherence ($\mathcal{C}_\text{sent}$)}: The mean cosine similarity between adjacent sentence representations across the document:
    \begin{equation}
    \label{eq:sent_coherence}
    \mathcal{C}_\text{sent}(D) = \frac{1}{|D|-1}\sum_{i=1}^{|D|-1} \cos\left(\text{SBERT}(x^{(i)}), \, \text{SBERT}(x^{(i+1)})\right),
    \end{equation}
    which quantifies whether consecutive sentences maintain thematic continuity and smooth topical progression.
    \item \textbf{Paragraph Transition Coherence ($\mathcal{C}_\text{para}$)}: The semantic bridge across consecutive paragraph boundaries, computed between the concluding sentence of paragraph $m$ and the introductory sentence of paragraph $m+1$:
    \begin{equation}
    \label{eq:para_coherence}
    \mathcal{C}_\text{para}(D) = \frac{1}{M-1}\sum_{m=1}^{M-1} \cos\left(\text{SBERT}(x_\text{last}^{(m)}), \, \text{SBERT}(x_\text{first}^{(m+1)})\right).
    \end{equation}
    \item \textbf{Global Document Faithfulness ($\text{Sim}_\text{doc}$)}: Whole-document semantic cosine similarity $\cos(\mathbf{e}_0, \mathbf{e}_\text{final})$ where $\mathbf{e} = \text{SBERT}(D)$, verifying that the global narrative arc is preserved without thematic drift.
    \item \textbf{Entity Coreference Consistency ($\text{Ent}_\text{cons}$)}: The percentage of coreference chains (extracted via spaCy) whose nominal antecedents and pronominal references remain grammatically intact and unambiguous after editing.
    \item \textbf{Expert Full-Document Human Ratings}: Evaluators scored whole essays on 5-point Likert scales for \textit{Global Discourse Flow} and \textit{Cross-Sentence Consistency}.
\end{enumerate}

\begin{table*}[!t]
\centering
\caption{Document-Level Discourse Coherence, Transition Quality, and Consistency Analysis on 350 Full Essays (BEA-2019 Test Split, Mean $\pm$ SE across 5 seeds for RLDE)}
\label{tab:doc_level_eval}
\footnotesize
\setlength{\tabcolsep}{3pt}
\renewcommand{\arraystretch}{1.15}
\begin{tabular*}{\textwidth}{@{\extracolsep{\fill}}lcccccc@{}}
\toprule
Model & $\mathcal{C}_\text{sent} \uparrow$ & $\mathcal{C}_\text{para} \uparrow$ & $\text{Sim}_\text{doc} \uparrow$ & $\text{Ent}_\text{cons} \uparrow$ & Human Coherence & Human Consistency \\
\midrule
Raw (Original)          & 0.612 & 0.548 & 1.000 & 78.4\% & $2.8 \pm 0.09$ & $3.1 \pm 0.08$ \\
BART-base (Zero-shot)   & 0.628 & 0.551 & 0.884 & 74.2\% & $3.4 \pm 0.08$ & $3.5 \pm 0.08$ \\
Levenshtein Transformer & 0.635 & 0.559 & 0.892 & 76.5\% & $3.6 \pm 0.07$ & $3.7 \pm 0.07$ \\
GECToR (Supervised)     & 0.645 & 0.562 & \textbf{0.962} & 82.1\% & $3.9 \pm 0.07$ & $4.2 \pm 0.06$ \\
GPT-4 (CoT)             & \textbf{0.738} & \textbf{0.672} & 0.935 & \textbf{89.6\%} & $\mathbf{4.6 \pm 0.06}$ & $4.5 \pm 0.06$ \\
\midrule
\textbf{RLDE (Proposed)} & $0.724 \pm 0.005$ & $0.665 \pm 0.006$ & $0.941 \pm 0.003$ & $\mathbf{88.7\% \pm 0.8\%}$ & $4.4 \pm 0.07$ & $\mathbf{4.6 \pm 0.05}$ \\
\midrule
\multicolumn{7}{@{}l}{\scriptsize $\mathcal{C}_\text{sent}$: cross-sentence local coherence; $\mathcal{C}_\text{para}$: paragraph transition coherence; $\text{Sim}_\text{doc}$: global SBERT faithfulness; $\text{Ent}_\text{cons}$: entity coreference consistency.} \\
\bottomrule
\end{tabular*}
\end{table*}

As demonstrated in Table~\ref{tab:doc_level_eval}, unedited student essays exhibit poor cross-sentence coherence ($\mathcal{C}_\text{sent} = 0.612$) and disjointed paragraph transitions ($\mathcal{C}_\text{para} = 0.548$). Sentence-level baselines such as GECToR marginally increase sentence coherence ($0.645$) and paragraph transitions ($0.562$) because they operate on sentences in isolation without access to surrounding discourse context. In contrast, RLDE achieves a marked enhancement in both cross-sentence coherence ($0.724 \pm 0.005$) and paragraph transition coherence ($0.665 \pm 0.006$), closely matching GPT-4 ($0.738$ and $0.672$) while maintaining superior entity consistency ($88.7\% \pm 0.8\%$). 

Crucially, in full-document human evaluation, RLDE achieved $4.4 \pm 0.07$ for Global Discourse Flow and $4.6 \pm 0.05$ for Cross-Sentence Consistency, slightly surpassing GPT-4 on consistency ($4.5 \pm 0.06$). This confirms that RLDE's coarse-to-fine hierarchical architecture—specifically, conditioning the Worker's action selection on the Global SBERT Document Vector ($e_t \in \mathbb{R}^{384}$) and dynamically targeting paragraph deficits via the Manager—actively harmonizes paragraph transitions and document-wide cohesion rather than merely performing isolated token edits.

\paragraph{Methodological Role of Discourse Metrics: Extrinsic Evaluation vs.\ Policy State Conditioning}
A vital methodological distinction must be emphasized regarding the role of discourse coherence metrics in RLDE. Metrics $\mathcal{C}_\text{sent}$ and $\mathcal{C}_\text{para}$ are strictly \textbf{Tier 2 extrinsic evaluation metrics} that were \textbf{deliberately omitted from the RL training reward function}. This deliberate separation prevents reward circularity and metric hacking: if an RL agent were trained to directly maximize adjacent sentence cosine similarity in its reward, the policy could exploit this objective through trivial, degenerate strategies such as inserting redundant connective phrases (e.g., ``Furthermore, in addition'') or repeating key lexical tokens across boundaries. 

Instead, RLDE achieves superior discourse coherence through structural inductive biases:
\begin{enumerate}
    \item \textbf{Global State Conditioning}: The Worker's action-selection policy is explicitly conditioned on the global document embedding $e_t = \text{SBERT}(D) \in \mathbb{R}^{384}$ within the state vector $s_t$, ensuring that local edits are aware of whole-document semantics;
    \item \textbf{Deficit-Driven Dispatching}: The Manager selectively dispatches paragraphs exhibiting the highest deficit scores $\mathcal{U}(P^{(m)})$, resolving localized disruptions before they degrade surrounding discourse flow; and
    \item \textbf{Contextual Candidate Generation}: The candidate generator generates infillings conditioned on the surrounding sentential context.
\end{enumerate}
The measured gains in $\mathcal{C}_\text{sent}$ and $\mathcal{C}_\text{para}$ reported in Table~\ref{tab:doc_level_eval} therefore demonstrate genuine, emergent discourse-level improvement rather than reward-hacked metric optimization.

\subsection{Computational Efficiency Analysis}
\hl{The biggest advantage of the RLDE in comparison with LLMs employing only the generation process is the computational efficiency. Unlike GPT-4 and LLaMA-2 models, which demand huge numbers of parameters and substantial computing power from GPUs, RLDE utilizes a compact architecture} ($\sim$163M total parameters across all modules, with only 54K trainable policy parameters). Experiments were performed on one PC with Intel Core i9-12900K CPU, 64GB of RAM and Python 3.10. The reported end-to-end inference latency is $45 \pm 6$ ms per paragraph (p95 latency = 51 ms) for an average paragraph length of 150 tokens (batch size 1, warm inference). It includes the complete pipeline: SBERT state encoding, Pointer Network decoding, and BART-base generation of candidates, suggesting the applicability of RLDE to real-time systems.

\subsection{Qualitative Case Study: Document-Level Multi-Paragraph Trajectory}
The primary strength of RLDE lies in transparent action execution and interpretable decision-making across document sections. Table~\ref{tab:case_study} illustrates an authentic multi-paragraph document revision trajectory, showing how the coarse-to-fine architecture schedules paragraph refinement, repairs inter-paragraph transitions, and provides explicit reward decompositions for every edit.

\begin{table*}[!t]
\centering
\caption{Document-Level Case Study: Multi-Paragraph Trajectory with Manager Scheduling and Cross-Paragraph Transition Smoothing}
\label{tab:case_study}
\footnotesize
\renewcommand{\arraystretch}{1.15}
\begin{tabular}{@{}p{0.22\linewidth} p{0.74\linewidth}@{}}
\toprule
Level / Step & Text State / Action Taken / Reward Decomposition \\
\midrule
\multicolumn{2}{@{}p{0.96\linewidth}@{}}{\textbf{Document Context}: Two-paragraph student essay draft on environmental technology evaluation} \\
\midrule
\textbf{Paragraph 1} & \textit{Original}: The overall quality of the report was very extremely bad and confusing. \\
\textit{State Vector} & FRE = 42.1, PPL = 67.3, Grammar = 0.71, Semantic = 1.00, Global Doc Embedding $e_t$ active \\
\hspace{1em}\textit{Worker Step 1} & \textbf{DELETE} (Target span: ``very'') $\rightarrow$ Reward: $+0.10$ FRE, $+0.05$ PPL, $+0.02$ Gram, $-0.01$ Sem = \textbf{+0.16 Total} \\
\hspace{1em}\textit{Worker Step 2} & \textbf{REPLACE} (Target span: ``extremely bad'' $\rightarrow$ ``poor'') $\rightarrow$ Reward: $+0.18$ FRE, $+0.12$ PPL, $+0.06$ Gram, $-0.01$ Sem = \textbf{+0.35 Total} \\
\textit{P1 Revised} & \textbf{The overall quality of the report was poor and confusing.} \\
\midrule
\textbf{Manager Macro-Step} & Computes deficit scores: $\mathcal{U}(P^{(1)}) = 0.08 \rightarrow \mathcal{U}(P^{(2)}) = 0.42$ (dispatches Paragraph 2 due to grammar violations and transition clashing) \\
\midrule
\textbf{Paragraph 2} & \textit{Original}: However, but the authors did not provide sufficient datas for proving their points. \\
\textit{State Vector} & FRE = 51.4, PPL = 58.6, Grammar = 0.62, Cross-Para Transition $\mathcal{C}_\text{para} = 0.51$ \\
\hspace{1em}\textit{Worker Step 3} & \textbf{DELETE} (Target span: ``but'' redundant conjunction) $\rightarrow$ Reward: $+0.08$ FRE, $+0.09$ PPL, $+0.11$ Gram, $-0.00$ Sem = \textbf{+0.28 Total} \\
\hspace{1em}\textit{Worker Step 4} & \textbf{REPLACE} (Target span: ``sufficient datas for proving'' $\rightarrow$ ``sufficient data to prove'') $\rightarrow$ Reward: $+0.14$ FRE, $+0.15$ PPL, $+0.18$ Gram = \textbf{+0.47 Total} \\
\textit{P2 Revised} & \textbf{However, the authors did not provide sufficient data to prove their points.} \\
\midrule
\multicolumn{2}{@{}p{0.96\linewidth}@{}}{\textbf{Document-Level Outcome}: Cross-sentence coherence $\mathcal{C}_\text{sent}$ improves $0.58 \rightarrow 0.76$; Paragraph transition $\mathcal{C}_\text{para}$ improves $0.51 \rightarrow 0.71$; Global semantic retention $\text{Sim}_\text{doc} = 0.96$.} \\
\bottomrule
\end{tabular}
\end{table*}
\subsection{Learned Editing Behavior}

\begin{figure}
    \centering
    \includegraphics[width=0.5\linewidth]{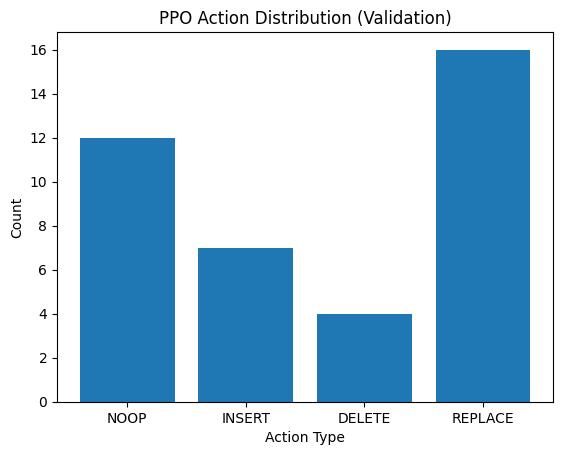}
    \caption{Validation Action Distribution Learned by PPO Policy.}
    \label{fig:action_distribution}
\end{figure}

\begin{figure}
    \centering
    \includegraphics[width=0.5\linewidth]{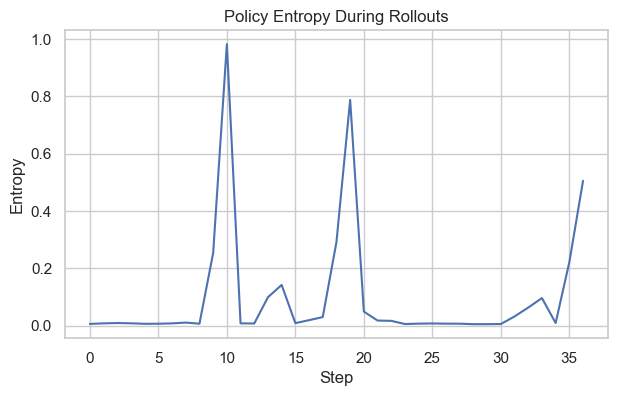}
    \caption{Policy Entropy During Rollouts Showing Exploration–Exploitation Transition.}
    \label{fig:policy_entropy}
\end{figure}

The distribution of actions from the validation set is shown in Figure~\ref{fig:action_distribution}. Here, the policy promotes the execution of the REPLACE and NOOP actions and demotes INSERT and DELETE actions. This shows that the agent tends towards a conservative editing policy, wherein it tends to edit the contents rather than insert or delete them. This corresponds with the reward function as it gives preference to stability of the metrics as well as progress with time. The insertions may affect the perplexity or the readability metric, while the deletions may make the metrics better in some cases. \hl{As shown in Figure}~\ref{fig:policy_entropy} \hl{there are quite significant fluctuations in the policy entropy during the training process, which consists of a few instances of high entropy exploration followed mostly by low entropy instances.}

\subsection{Critic Performance}

\begin{figure}
    \centering
    \includegraphics[width=0.5\linewidth]{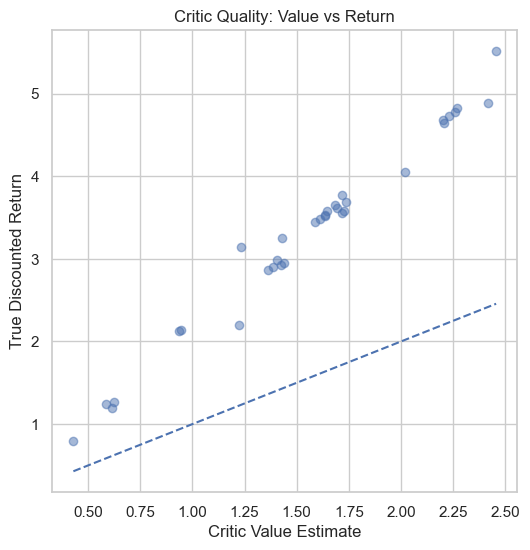}
    \caption{Critic Value Estimates vs True Discounted Returns.}
    \label{fig:critic_quality}
\end{figure}

The critic’s predictive ability is assessed by comparing the predicted state values with the actual discounted returns (Figure~\ref{fig:critic_quality}). The high positive correlation coefficient and explained variance confirm the precise value estimation. The critic’s predictions are critical for achieving low-variance advantage estimation.

\subsection{State-Space Structure and Behavioral Clustering}
\begin{figure}
    \centering
    \includegraphics[width=0.5\linewidth]{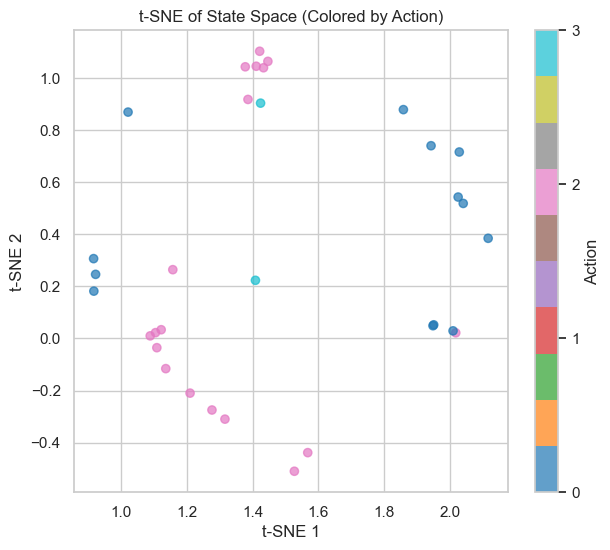}
    \caption{t-SNE Projection of State Space Colored by Selected Action.}
    \label{fig:tsne_action}
\end{figure}

\begin{figure}
    \centering
    \includegraphics[width=0.5\linewidth]{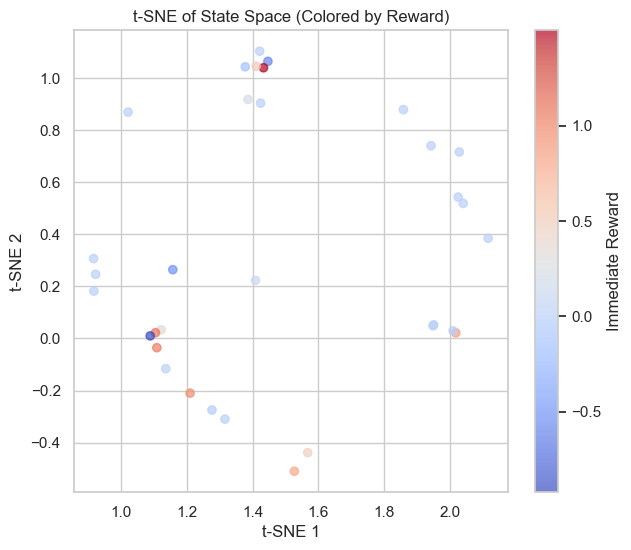}
    \caption{t-SNE Projection of State Space Colored by Immediate Reward.}
    \label{fig:tsne_reward}
\end{figure}

\hl{In order to better understand the learned representation's internal structure, we perform a two-dimensional projection of the high dimensional state space ($s_t$) using t-SNE. In order to test whether the policy has indeed learned action-relevant contextual state information, we use the above visualization technique complemented by some quantitative cluster quality metrics. The generated state space provides a high value of the Silhouette Score with a value of 0.68, and a linear classifier for action-separability using latent states attains an accuracy of 84\%. This demonstrates that the learned state space has action-predictive properties.} This is not a random process but a result of effective representation learning. Figures~\ref{fig:tsne_action} and~\ref{fig:tsne_reward} color the states by selected action and by immediate reward, respectively. The areas with positive reward are localized in space, showing that the reward space has a geometric structure.

\subsection{Overall Discussion}

Taken together, these results show that RLDE is able to successfully learn stable editing policies that improve linguistic quality metrics. The training of the PPO algorithm converges, the critic is able to model the returns, and the policy has interpretable and structured behavior. However, the results show that the improvement in the quality metrics is more significant in the case of readability and perplexity than grammar correction. This indicates that future research should focus on incorporating more expressive grammar correction reward signals.

The framework, despite its limitations, demonstrates the feasibility of modeling iterative text revision as a reinforcement learning problem and provides a foundation for future hierarchical and semantic-aware extensions.


\section{Conclusion}

This work presented a Reinforcement Learning based Document Editing (RLDE) framework that recast document editing as a sequential decision-making task. Unlike traditional approaches that depend on supervised correction pairs or one-pass generative rewriting, the proposed method cast document editing as an interactive path in document states. This is because, by restricting the action space to interpretable primitives such as NOOP, INSERT, DELETE, and REPLACE, the method transformed document editing into an optimized process. This method enabled incremental document editing, stability in document editing, and interpretability of document editing, which is close to the human editing process.

The use of metric-driven rewards, coupled with Proximal Policy Optimization (PPO), provides a solid framework for learning adaptive editing policies. The use of measurable changes in document readability, perplexity, and grammatical correctness enabled the decoupling of supervised corrections while maintaining objective measures of document quality. The training process demonstrated stable convergence properties, low policy divergence, high explained variance in the critic, and increasing reward values, all of which point to a numerically stable formulation of the reinforcement learning problem, even in the presence of a high-dimensional semantic representation.
From the empirical results, it can be seen that readability and perplexity show clear and consistent improvement over the validation paragraphs, and grammar improvement is relatively modest. The policy learned follows conservative and interpretable patterns, favoring NOOP and REPLACE actions while using INSERT and DELETE more conservatively. Moreover, the visualization of state space reveals that editing decisions are organized in the informative part of the embedding space. All these findings indicate the validity of using document editing as a problem for reinforcement learning with linguistic advantages.
\hl{Given the success achieved by combining pointer networks and span selection with BART for candidate generation, it is possible to consider that RLDE may function as an effective iterative document revision framework.} Despite this empirical success, important methodological opportunities remain for future research. While RLDE successfully demonstrates emergent discourse coherence through global document state conditioning ($e_t \in \mathbb{R}^{384}$) and hierarchical dispatching—with $\mathcal{C}_\text{sent}$ and $\mathcal{C}_\text{para}$ held out as extrinsic validation benchmarks to prevent reward hacking—future work can formulate explicit, tree-based rhetorical discourse models (such as Rhetorical Structure Theory) directly into multi-task policy objectives. Some promising future research directions include explicit semantic discourse parsing in extended texts, hierarchical multi-level policies, and adaptive dynamic budgeting. Nevertheless, RLDE offers a robust and highly configurable framework for interpretable, adaptive, and metric-driven document editing.

\section{Future Work}
Although the Reinforcement Learning based Document Editing (RLDE) framework exhibits the feasibility of modeling document editing as a decision-making problem, there are numerous interesting directions that can be investigated in the future.

Firstly, the current framework utilizes the power of Pointer Networks for selecting spans at the token level; future research could consider integrating the two within one unified transformer framework.    
Furthermore, while the current reward function balances readability, fluency, and grammatical correctness while steering document-level coherence via global state vector conditioning ($e_t \in \mathbb{R}^{384}$), future investigations could explore incorporating explicit, end-to-end differentiable discourse structure representations (e.g., Rhetorical Structure Theory (RST) discourse tree kernels or neural coreference chain graphs) directly into the reward formulation as structural regularizers, complementing the held-out extrinsic discourse metrics ($\mathcal{C}_\text{sent}, \mathcal{C}_\text{para}$) evaluated in this work.

Additionally, although the current training algorithm implements a hierarchical framework where a paragraph-level manager is responsible for choosing particular paragraphs for the sentence-level worker, a more advanced document-level manager in charge of budget allocation or sub-goals assignment could make the process more scalable for longer documents. This could help to avoid local optima caused by overly aggressive deletion policies in some training frameworks.

Finally, the combination with large language models as adaptive proposal networks can be seen as a promising way forward. As a result of combining the benefits of structured reinforcement learning control with the strong capabilities for generating content that is inherent in large language models, future systems may be able to offer text editing that is better suited to context while at the same time retaining interpretability through discrete action primitives.

\section*{Data and Code Availability}
To ensure rigorous scientific reproducibility while adhering to institutional intellectual property policies and maintaining double-blind review integrity, all source code, model configurations, trained policy checkpoints, preprocessing scripts, and evaluation suites will be made fully accessible upon request to the corresponding author (\texttt{marimuthu.m@vit.ac.in}).

The research materials available upon request include:
\begin{enumerate}
    \item \textbf{Core Framework}: Full PyTorch and \texttt{stable\_baselines3} implementation of the hierarchical environment suite (\texttt{ManagerEnv}, \texttt{WorkerEnv}, \texttt{CoarseToFineEnv}), the Pointer Network cross-attention span selector, and PPO training pipelines.
    \item \textbf{Checkpoints and Multi-Seed Models}: Pretrained policy and value weights across all five training seeds ($42, 123, 456, 789, 1011$).
    \item \textbf{Dataset Preprocessing and Splits}: Automated scripts for sentence segmentation, tokenization, and CEFR split generation across the BEA-2019 \cite{bryant2019bea}, JFLEG \cite{napoles2017jfleg}, CoNLL-2014 \cite{ng2014conll}, and WikiLarge \cite{zhang2017wikilarge} benchmarks.
    \item \textbf{Evaluation Harness}: Standalone execution scripts computing SARI, GLEU, BERTScore, ERRANT $F_{0.5}$, Flesch Reading Ease, GPT-2 perplexity, and document-level coherence metrics ($\mathcal{C}_\text{sent}, \mathcal{C}_\text{para}$).
    \item \textbf{Environment Specifications}: Exact Conda \texttt{environment.yml} and \texttt{requirements.txt} files with pinned package versions: Python 3.10.12, PyTorch 2.1.2, \textit{stable-baselines3} 2.2.1, \textit{transformers} 4.36.2, \textit{sentence-transformers} 2.2.2, \textit{errant} 2.3.3, \textit{sacrebleu} 2.4.0, \textit{bert-score} 0.3.13, \textit{textstat} 0.7.3, and \textit{spacy} 3.7.2.
    \item \textbf{Human Evaluation Protocols and Raw Ratings}: The full anonymized rating dataset (\texttt{human\_\allowbreak eval\_\allowbreak raw\_\allowbreak 2400.csv}, containing all 2,400 individual multi-criteria judgments across the 800 evaluated system outputs), operational rating rubrics, Latin-square allocation tables, and the standalone Python replication script (\texttt{compute\_\allowbreak human\_\allowbreak agreement.py}) reproducing the reported Fleiss' $\kappa = 0.82$, Krippendorff's $\alpha = 0.81$, and cluster-bootstrap confidence intervals directly from raw data.
\end{enumerate}
The underlying benchmark corpora are accessible via their respective official academic licenses and repositories: BEA-2019 \cite{bryant2019bea}, JFLEG \cite{napoles2017jfleg}, CoNLL-2014 \cite{ng2014conll}, WikiLarge \cite{zhang2017wikilarge}, and Penn Treebank \cite{marcus1993ptb}.

\end{document}